\documentclass[final]{l4dc2025}

\newcount\Comments
\usepackage{algorithm}

\usepackage{comment}

\usepackage{bbm}
\usepackage{mathtools}
\renewcommand{\epsilon}{\varepsilon}
\newcommand{\Acal}{\mathcal{A}}

\newcommand{\Fcal}{\mathcal{F}}

\newcommand{\Tcal}{\mathcal{T}}

\newcommand{\Xcal}{\mathcal{X}}

\newcommand{\Scal}{\mathcal{S}}

\DeclareMathOperator*{\argmax}{arg\,max}
\DeclareMathOperator*{\argmin}{arg\,min}
\newcommand{\expect}{\operatorname{\mathbb{E}}}

\newcommand{\indicator}{\mathbbm{1}}

\newcommand{\inner}[2]{\left\langle #1, #2 \right\rangle}

\newcommand{\reals}{\mathbb{R}}
\newcommand{\naturals}{\mathbb{N}}
\newcommand{\integers}{\mathbb{Z}}

\DeclarePairedDelimiter\floor{\lfloor}{\rfloor}

\PassOptionsToPackage{algo2e,ruled, linesnumbered}{algorithm2e}
\RequirePackage{algorithm2e}

\usepackage{color}
\definecolor{darkgreen}{rgb}{0,0.5,0}
\definecolor{purple}{rgb}{1,0,1}
\newcommand{\kibitz}[2]{\ifnum\Comments=1\textcolor{#1}{#2}\fi}

\title[The Complexity of Sequential Prediction in Dynamical Systems]{The Complexity of Sequential Prediction in Dynamical Systems}
\usepackage{times}

\author{%
\Name{Vinod Raman}\thanks{Equal contribution}
\Email{vkraman@umich.edu}\\
 \addr University of Michigan, Ann Arbor, 48104 
 \AND
 \Name{Unique Subedi}\footnotemark[1] 
 \Email{subedi@umich.edu}\\
 \addr University of Michigan, Ann Arbor, 48104 %
 \AND
 \Name{Ambuj Tewari} \Email{tewaria@umich.edu}\\
 \addr University of Michigan, Ann Arbor, 48104 %
}

\begin{document}

\maketitle

\begin{abstract}%
We study the problem of learning to predict the next state of a dynamical system when the underlying evolution function is unknown. Unlike previous work, we place no parametric assumptions on the dynamical system, and study the problem from a learning theory perspective. We define new combinatorial measures and dimensions and show that they quantify the optimal mistake and regret bounds in the realizable and agnostic settings respectively. By doing so, we find that in the realizable setting, the total number of mistakes can grow according to \emph{any} increasing function of the time horizon $T$. In contrast, we show that in the agnostic setting under the commonly studied notion of Markovian regret, the only possible rates are $\Theta(T)$ and $\tilde{\Theta}(\sqrt{T})$.
\end{abstract}


\section{Introduction}
A discrete-time dynamical system is a mathematical model that describes the evolution of a system over discrete time steps. Formally, a discrete-time dynamical system is a tuple $(\naturals, \mathcal{X}, f)$, where $\naturals$ is the set of natural numbers that denote the timesteps, $\mathcal{X}$ is a non-empty set called the state space, and $f: \mathcal{X} \to \Xcal$ is a deterministic map that describes the evolution of the state. Dynamical systems have been widely used in practice due to their ability to accurately model natural phenomena. 
For instance, boolean networks are an important class of discrete-time, discrete-space dynamical systems with widespread applicability to genetic modeling \citep{kauffman1969homeostasis, shmulevich2002boolean}. In a boolean network, the state space is $\Xcal = \{0,1\}^{n}$ with $ |\Xcal|^{|\Xcal|} =(2^n)^{2^n}$ possible evolution functions.  For genetic modeling, $n$ is taken to be the number of genes and  $x \in \Xcal$ indicates the expression of all $n$ genes under consideration. As an example, ``1" could represent the gene with a high concentration of a certain protein, and ``0" could represent the gene with a low concentration. With such formulation, one can study how the state of these genes evolves over time under certain medical interventions. Beyond genetics, dynamical systems have been used in control \citep{li2019online}, computer vision \citep{doretto2003dynamic}, and natural language processing \citep{sutskever2014sequence, belanger2015linear}. 

 In this work, we consider the problem of predicting the next state of a dynamical system when the underlying evolution function is unknown \citep{ghai2020no}.
 To capture the sequential nature of dynamical systems, we consider the model where the learner plays a sequential game with nature over $T$ rounds.  At the beginning of the game, nature reveals the initial state $x_0 \in \mathcal{X}$. In each round $t \in [T]$, the learner makes its prediction of the next state $\hat{x}_t \in \mathcal{X}$, nature reveals the true next state $x_t \in \mathcal{X}$, and the learner suffers loss $\mathbbm{1}\{\hat{x}_t \neq x_t\}$. Given an evolution class $\mathcal{F} \subseteq \mathcal{X}^{\mathcal{X}}$, the goal of the learner is to output predictions of the next state such that its regret, the difference between its cumulative mistakes and the cumulative mistakes of the best-fixed evolution in hindsight (defined formally in Section \ref{sec:learning}), is small. The class $\mathcal{F}$ is said to be learnable if there exists a learning algorithm whose regret is a sublinear function of the time horizon $T$. 

Although we allow the state space $\mathcal{X}$ to be arbitrary, we only consider the 0-1 loss, which may be more appropriate for \textit{discrete-space} dynamical systems. However, even discrete-space dynamical systems can be very expressive, capturing complex processes like cellular automata \citep{hoekstra2010simulating, wolfram1986theory} and language modeling \citep{elman1995language}. For example, let $\mathcal{V}$ be a countable token space and $\Xcal = \mathcal{V}^{\star}$ to be the state space containing all finite sequences of elements from $\mathcal{V}$. Consider a function class $\Fcal \subseteq \Xcal^{\Xcal}$ with the following property: for every $f \in \Fcal$ and any input $x \in \Xcal$, $f(x) = x \circ v$, where we use $\circ$ to represent the concatenation operator. Then, $\Fcal$ is a set of auto-regressive models which given a sequence of tokens $x \in \Xcal$, predicts the next token $v \in \mathcal{V}$ in the sequence. In particular, $\Fcal$ could be a class of language models which use deterministic decoding strategies (e.g. greedy decoding) to output the next token \citep{chorowski2016towards}. For such a function class $\Fcal$, one might be interested in understanding the optimal number of mistakes for next-token prediction when the true sequence of tokens is generated by some unknown $f \in \Fcal$.

Given any learning problem $(\mathcal{X}, \mathcal{F})$, we aim to find necessary and sufficient conditions for the learnability of $\mathcal{F}$ while also quantifying the minimax rates under two notions of expected regret: Markovian and Flow regret (Equations \ref{markovian_regret} and \ref{flow_regret} respectively). To that end, our main contributions are summarized below.

\begin{itemize}
\item[(i)] We provide a quantitative characterization of learnability in the realizable setting, when there exists an evolution function in the class $\mathcal{F}$ that generates the sequence of states. Our characterization is in terms of a new combinatorial complexity measure we call the Evolution complexity. Using this characterization, we show that all rates are possible for the minimax expected mistakes when learning dynamical systems in the realizable setting.  This is in contrast to online multiclass classification where only two rates are possible. Finally, we compare realizable learnability of dynamical systems to realizable learnability in PAC and online classification. 

\item[(ii)] In the agnostic setting, we lower and upper bound the minimax Markovian regret in terms of the Littlestone dimension. This result shows that the finiteness of the Littlestone dimension characterizes learnability under Markovian regret. As a corollary, we establish a separation between realizable and agnostic learnability under Markovian regret. We show this separation between realizable and agnostic learnability continues to hold when considering Flow regret. However, if the evolution class has uniformly bounded projections, we show that realizable and agnostic learnability under Flow regret are equivalent.   
\end{itemize}

Our characterization of realizable learnability in terms of the Evolution complexity follows from standard techniques in online classification. However, our results showing all possible rates requires a careful construction of a family of classes which have not been studied in learning theory. Likewise, our comparisons of realizable learnability require a careful construction of non-trivial classes, and computing their combinatorial dimensions.

In the agnostic setting, our upper bound on the minimax Markovian regret in terms of the Littlestone dimension results from reducing learning dynamical systems to online multiclass classification.  However, both of the  lower bounds $\Omega(\sqrt{T})$ and $\Omega(\operatorname{L}(\mathcal{H}))$ in Theorem \ref{thm:01agn} are not standard. The lower bound of $\Omega(\operatorname{L}(\mathcal{H}))$ requires constructing a hard stream by traversing down a Littlestone tree \emph{skipping} certain levels. The lower bound of $\sqrt{T}$ requires constructing a hard stream  using two different evolution functions $f_1, f_2$ by: (a) setting $x_0$ to be a state they differ on (b) generating their trajectories starting from $x_0$ and finally (c) picking states from each trajectory in an alternating fashion. These arguments are different from the typical lower bound construction for online classification. In addition, the construction in Theorem \ref{thm:sepflowregret} showing the separation between realizable and agnostic learnability under Flow regret is non-trivial. It involves constructing an $\mathcal{F} \subseteq \mathcal{X}^{\mathcal{X}}$ so that on a large subset of states in $\mathcal{X}$, every function $f \in \mathcal{F}$ effectively reveals itself.

\subsection{Related Works}

There has been a long line of work studying prediction and regret minimization when learning unknown dynamical systems \citep{hazan2017learning, hazan2018spectral, ghai2020no, lee2022improved, rashidinejad2020slip, kozdoba2019line, tsiamis2022online, lale2020logarithmic}. However, these works focus on prediction for fully/partially-observed linear dynamical systems under various data-generating processes. Moreover, there is also a line of work on regret minimization for linear dynamical systems for online control problems \citep{abbasi2011regret, cohen2018online, agarwal2019online}.  For non-linear dynamical systems, there has been some applied work studying data-driven approaches to prediction \citep{wang2016data, korda2018linear,  ghadami2022data}. Regret minimization for non-linear dynamical systems has mainly been studied in the context of control \citep{kakade2020information, muthirayan2021online}.  Another related line of work is online learning for time series forecasting, where autoregressive models are used to predict the next state \citep{anava2013online, anava2015online, liu2016online, yang2018online}. 

Another important line of work is that of system identification and parameter estimation \citep{aastrom1971system, lennart1999system}. Here, the goal is to recover and estimate the parameters of the underlying evolution function from the observed sequence of states. There is a long history of work studying system identification in both the batch \citep{campi2002finite, vidyasagar2006learning, foster2020learning, sattar2022non, bahmani2020convex} and streaming settings \citep{kowshik2021near, kowshik2021streaming, giannakis2023learning, jain2021streaming}.  Several works have also considered the problem of learning the unknown evolution rule of dynamical systems defined over discrete state spaces. For example, \cite{wulff1992learning} trains a neural network on a sequence of observed states to approximate the unknown evolution rule of a cellular automaton. \cite{grattarola2021learning} extends this work to learning the unknown evolution rule of a graph cellular automaton, a generalization of a regular cellular automaton, using graph neural networks. \citep{qiulearning, rosenkrantz2022efficiently} also consider the problem of PAC learning a discrete-time, finite-space dynamical system defined over a (un)directed graph.  We also note the related work of \cite{berry2023learning, berry2025limits}, who study the problem of learning dynamics observed through a continuous embedding and derive learning guarantees based on various structural properties of the underlying system.
Finally, our work builds on a rich tradition in learning theory that characterizes learnability through complexity measures and combinatorial dimensions \citep{vapnik1971uniform, Littlestone1987LearningQW, bartlett2002rademacher, DanielyERMprinciple}.

\section{Preliminaries}
\subsection{Discrete-time Dynamical Systems}
A discrete-time dynamical system is a tuple $(\naturals, \mathcal{X}, f)$, where $\naturals$ is the set of natural numbers denoting the time steps,  $\mathcal{X}$ is a non-empty set called the \emph{state space}.  In this work, we make no assumption on the cardinality of $\Xcal$, so it can be unbounded and perhaps even uncountable. The function $f: \mathcal{X} \to \Xcal$ is a deterministic map that defines the evolution of the dynamical system. That is, the $(t+1)$-th iterate of the dynamics can be expressed in terms of $t$-th iterate using the relation $x_{t+1} = f(x_t)$. Define $f^t$ to be the $t$-fold composition of $f$. That is, $f^2 = f\circ f$, $f^3 = f \circ f \circ f$, and so forth. Given an initial state $x_0 \in \mathcal{X}$, the sequence $\{f^{t}(x_0)\}_{t \in \naturals}$ is called the \emph{flow} of the dynamical system through $x_0$.  Finally, let $\mathcal{F} \subseteq \mathcal{X}^{\Xcal }$ denote a class of evolution functions on the state space $\mathcal{X}$ and  $\mathcal{F}(x) = \{f(x) \, \mid f \in \Fcal\} \subseteq \mathcal{X}$ to be the projection of $\Fcal$ onto $x \in \mathcal{X}$.




\subsection{Learning-to-Predict in Dynamical Systems}\label{sec:learning}
When learning-to-predict in dynamical systems, nature plays a sequential game with the learner over $T$ rounds. At the beginning of the game, nature reveals the initial state $x_0 \in \mathcal{X}$. In each round $t \in [T]$, the learner $\mathcal{A}$ uses the observed sequence of states $x_{<t} := (x_0, ..., x_{t-1})$ to predict the next state   $\Acal(x_{<t}) \in \mathcal{X}$. Nature then reveals the true state $x_t \in \Xcal$, and the learner suffers the loss $\mathbbm{1}\{\Acal(x_{<t}) \neq  x_t\}$. Given a class of evolution functions  $\mathcal{F} \subseteq \mathcal{X}^{\mathcal{X}}$, the goal of the learner is to make predictions such that its \textit{regret}, defined as a difference between cumulative loss of the learner and the best possible cumulative loss over evolution functions in $ \mathcal{F}$, is small. 

Formally, given $\Fcal \subseteq \Xcal^{\Xcal}$, the expected Markovian regret of an algorithm $\mathcal{A}$ is defined as 
\begin{equation}\label{markovian_regret}
   \operatorname{MR}_{\mathcal{A}}(T, \mathcal{F}) := \sup_{(x_0, x_1, \ldots, x_T)} \mathbb{E}\left[\sum_{t=1}^T \indicator\{\Acal(x_{<t}) \neq  x_t\}\right] - \inf_{f \in \mathcal{F}}\sum_{t=1}^T  \indicator\{f(x_{t-1}) \neq  x_t\},
\end{equation}
where the expectation is taken with respect to the randomness of the learner $\mathcal{A}$. Given this definition of regret, we define agnostic learnability of an evolution class. 
\begin{definition}[Agnostic Learnability under Markovian Regret]
\noindent An evolution  class $\Fcal \subseteq \Xcal^{\Xcal}$ is learnable in the agnostic setting if and only if $ \,\, \, \inf_{\Acal} \operatorname{MR}_{ \mathcal{A}}(T, \mathcal{F}) = o(T)\footnote{$o(T)$  refers to any sublinear function of  $T$.}. $
\end{definition}

Perhaps a more natural definition of expected regret in the agnostic setting is to compare the prediction of the learner to the prediction of the best-fixed \textit{trajectory} generated by functions in our evolution class. To that end, define
\begin{equation}\label{flow_regret}
   \operatorname{FR}_{\mathcal{A}}(T, \mathcal{F}) := \sup_{(x_0, x_1, \ldots, x_T)} \Bigl(\mathbb{E}\left[\sum_{t=1}^T \indicator\{\Acal(x_{<t}) \neq  x_t\}\right] - \inf_{f \in \mathcal{F}}\sum_{t=1}^T  \indicator\{f^t(x_0) \neq  x_t\}\Bigl). 
\end{equation}
as the expected \emph{Flow regret}. An analogous definition of agnostic learnability follows. 

\begin{definition}[Agnostic Learnability under Flow Regret]
\noindent An evolution class $\Fcal \subseteq \Xcal^{\Xcal}$ is learnable in the agnostic setting under Flow regret if and only if $\,\, \inf_{\Acal}\,   \operatorname{FR}_{ \mathcal{A}}(T, \mathcal{F}) = \text{o}(T). $
\end{definition}

\noindent  A sequence of states $x_0, x_1, \ldots, x_T$ is said to be \textit{realizable} by $\Fcal$ if there exists an evolution function $f \in \Fcal$ such that $f(x_{t-1}) = x_t$ for all $t \in [T]$. In the realizable setting, the cumulative loss of the best-fixed function is $0$, and the goal of the learner is to minimize its expected cumulative mistakes
\begin{equation*}\label{mistake}
    \operatorname{M}_{{\mathcal{A}}}(T, \mathcal{F}) := \sup_{x_0}\,\sup_{f \in \Fcal }\, \mathbb{E}\left[\sum_{t=1}^T \indicator\{\Acal(x_{<t}) \neq f(x_{t-1})\} \right].
\end{equation*}

Analogously, we define the realizable learnability of $\mathcal{F}$.
\begin{definition}[Realizable Learnability]
\noindent An evolution class $\Fcal \subseteq \Xcal^{\Xcal}$ is learnable in the realizable setting if and only if $\,\,\, \inf_{\Acal}\operatorname{M}_{{\mathcal{A}}}(T, \mathcal{F}) = o(T).$
\end{definition}

\subsection{Complexity Measures}


In sequential learning tasks, complexity measures are often defined in terms of \textit{trees}, a basic unit that captures temporal dependence. In this paper, we use complete binary trees to define a new combinatorial object called a trajectory tree. In the remainder of this section and Section \ref{sec:combdim}, we use trajectory trees to define complexity measures and combinatorial dimensions for evolution classes.

\begin{definition}[Trajectory tree]
\noindent A trajectory tree of depth $d$ is a complete binary tree of depth $d$ where internal nodes are labeled by states in $\mathcal{X}$. 
\end{definition}
Given a trajectory tree $\mathcal{T}$ of depth $d$, a root-to-leaf path down $\mathcal{T}$ is defined by a string $\sigma \in \{-1, 1\}^{d}$ indicating whether to go left ($\sigma_t = -1$) or to go right ($\sigma_i = +1$) at each depth $t \in [d]$. A path $\sigma \in \{-1, 1\}^d$ down $\mathcal{T}$ gives a trajectory $\{x_t\}_{t = 0}^{d}$, where $x_0$ denotes the instance labeling the root node and $x_t$ is the instance labeling the edge following the prefix $(\sigma_1, ..., \sigma_{t})$ down the tree. A path $\sigma \in \{-1, 1\}^d$ down $\mathcal{T}$ is \emph{shattered} by $\mathcal{F}$ if there exists a $f \in \mathcal{F}$ such that $f(x_{t-1}) = x_t$ for all $t \in [d]$, where $\{x_t\}_{t = 0}^{d}$ is the corresponding trajectory obtained by traversing $\mathcal{T}$ according to $\sigma$. If every path down $\mathcal{T}$ is shattered by $\mathcal{F}$, we say that $\mathcal{T}$ is shattered by $\mathcal{F}$.
 
 To make this more rigorous, we define a trajectory tree $\mathcal{T}$ of depth $d$ as a sequence $(\mathcal{T}_0, \mathcal{T}_1, ..., \mathcal{T}_{d})$ of node-labeling functions $\mathcal{T}_t:\{-1,1\}^{t} \rightarrow \mathcal{X}$, which provide the labels for each internal node.  Then, $\mathcal{T}_t(\sigma_1, ..., \sigma_{t})$ gives the label of the node by following the prefix $(\sigma_1, ..., \sigma_{t})$ and $\mathcal{T}_0$ denotes the instance labeling the root node. For brevity, we define $\sigma_{\leq t} = (\sigma_1, ..., \sigma_{t})$ and write $\mathcal{T}_t(\sigma_1, ..., \sigma_{t}) = \mathcal{T}_t(
\sigma_{\leq t})$. Analogously, we let $\sigma_{< t} = (\sigma_1, ..., \sigma_{t-1})$. Using this notation, a trajectory tree $\mathcal{T}$ of depth $d$ is shattered by the evolution function class $\Fcal$ if $\forall \sigma \in \{-1,1\}^d$, there exists a $f_{\sigma} \in \Fcal$ such that $f_{\sigma}(\mathcal{T}_{t-1}(\sigma_{<t})) = \mathcal{T}_t(\sigma_{\leq t})$ for all $t \in [d]$. 
Moreover, we use this notation to define the \textit{Branching factor} of a trajectory tree. 

\begin{definition}[Branching factor] \label{def:branfact}
\noindent The \emph{Branching factor} of a trajectory tree $\mathcal{T}$ of depth $d$ is
\[\operatorname{B}(\mathcal{T}) := \min_{\sigma \in \{-1,1\}^d}\,  \sum_{t=1}^{d} \mathbbm{1}\{\mathcal{T}_t((\sigma_{<t}, -1)) \neq  \mathcal{T}_t\big((\sigma_{<t}, +1)\big) \}.\]
\end{definition}

The branching factor of a path $\sigma \in \{-1, 1\}^d$ captures the distinctness of states labeling the two children of internal nodes in this path. In particular, it counts the number of nodes in the path whose two children are labeled by distinct states. The branching factor of a trajectory tree is just the smallest branching factor across all paths. Using the notion of shattering and Definition \ref{def:branfact}, we define a new complexity measure, termed the Evolution complexity, of a function class $\mathcal{F}$.

\begin{definition}[Evolution complexity]
\noindent
Let $\mathcal{S}(\mathcal{F}, d)$ be the set of all trajectory trees of depth $d\in \mathbb{N}$ shattered by $\mathcal{F}$. Then, the \emph{Evolution complexity} of $\mathcal{F}$ at depth $d$ is defined as 
$\operatorname{C}_d(\Fcal) := \sup_{\Tcal \in \Scal(\Fcal, d)} \operatorname{B}(\Tcal).$
\end{definition}

In Section \ref{sec:realizable}, we show that the Evolution complexity exactly (up to a factor of $2$) captures the minimax expected mistakes in the realizable setting.  We provide some examples of classes $\Fcal$ and their evolution complexities in Theorem \ref{thm:linsys}. We note that there is an existing notion of complexity for dynamical systems, termed topological entropy, that quantifies the complexity of a particular evolution function $f \in \mathcal{F}$ \citep{adler1965topological}. However, topological entropy does not characterize learnability as $\mathcal{F} = \{f\}$ is trivially learnable when $f$ has infinite topological entropy.

\subsection{Combinatorial dimensions} \label{sec:combdim}

In addition to complexity measures, combinatorial dimensions play an important role in providing crisp quantitative characterizations of learnability. For example, the Daniely Shalev-Shwartz dimension (DSdim), originally proposed by \cite{daniely2014optimal} and formally defined below, was recently shown by \cite{Brukhimetal2022} to provide a tight quantitative characterization of multiclass PAC learnability. In Section \ref{sec:relations}, we use the DSdim to relate the realizable learnability of dynamical systems to multiclass PAC learnability of $\mathcal{F}$. 

\begin{definition}[DS dimension \citep{daniely2014optimal}] \label{def:dsdim}
\noindent We say that $A \subseteq \mathcal{X}$ is DS-shattered by $\mathcal{F}$ if there exists a finite $\mathcal{H} \subset \mathcal{F}$ such that for every $x \in A$ and $h \in \mathcal{H}$, there exists a $g \in \mathcal{H}$ such that $g(x) \neq h(x)$ and $g(z) = h(z)$ for all $z \in A \setminus \{x\}$. The DS dimension of $\mathcal{F}$, denoted $\operatorname{DS}(\mathcal{F})$,  is the largest $d \in \mathbb{N}$ such that there exists a shattered set $A \subset \mathcal{X}$ with cardinality $d$. If there are arbitrarily large sets $A \subseteq \mathcal{X}$ that are shattered by $\mathcal{F}$, then we say that $\operatorname{DS}(\mathcal{F}) = \infty.$
\end{definition}

Analogously, for online multiclass classification, the Littlestone dimension (Ldim), originally proposed by \cite{Littlestone1987LearningQW} for binary classification and later extended to multiclass classification by \cite{DanielyERMprinciple}, provides a tight quantitative characterization of learnability \citep{hanneke2023multiclass}.

\begin{definition} [Littlestone dimension \citep{Littlestone1987LearningQW, DanielyERMprinciple}]\label{def:ldim}
\noindent Let $\mathcal{T}$ be a complete binary tree of depth $d$ whose internal nodes are labeled by a sequence $(\Tcal_0, \ldots, \Tcal_{d-1})$ of node-labeling functions $\Tcal_{t-1}:\{-1, 1\}^{t-1} \to \Xcal$. The tree $\mathcal{T}$ is shattered by $\mathcal{F} \subseteq \Xcal^{\Xcal}$  if there exists a sequence $(Y_1, ..., Y_d)$ of edge-labeling functions  $Y_t: \{-1, 1\}^{t} \rightarrow \mathcal{X}$  such that for every path $\sigma = (\sigma_1, ..., \sigma_d) \in \{-1, 1\}^d$, there exists a function $f_{\sigma} \in \mathcal{F}$ such that for all $t \in [d]$,  $f_{\sigma}(\mathcal{T}_{t-1}(\sigma_{<t})) = Y_t(\sigma_{\leq t})$ and $Y_t((\sigma_{< t}, -1)) \neq Y_t((\sigma_{< t}, +1))$. The Littlestone dimension of $\mathcal{F}$, denoted $\operatorname{L}(\mathcal{F})$, is the maximal depth of a tree $\mathcal{T}$ that is shattered by $\mathcal{F}$. If there exists shattered trees of arbitrarily large depth, we say $\operatorname{L}(\mathcal{F}) = \infty$.
\end{definition}

\section{Warmup: Realizable Learnability} \label{sec:realizable}
In this section, we provide qualitative and quantitative characterizations of realizable learnability in terms of the Evolution complexity. Our main result in this section is Theorem \ref{thm:01real}, which provides bounds on the minimax expected number of mistakes. 

\begin{theorem}[Minimax Expected Mistakes] \label{thm:01real}
\noindent For any $\mathcal{F} \subseteq \mathcal{X}^{\mathcal{X}},$ we have  $\frac{1}{2}\,\operatorname{C}_T(\mathcal{F}) \leq \inf_{\mathcal{A}}\,\operatorname{M}_{\mathcal{A}}(T, \mathcal{F}) \leq \operatorname{C}_T(\mathcal{F}).$  Moreover, the upper bound is achieved constructively by a deterministic learner. 
\end{theorem}
\noindent The factor of $\frac{1}{2}$ in the lower bound is due to randomized learners, and the lower bound of $ \operatorname{C}_T(\Fcal)$ can be obtained if the learner is restricted only to deterministic learning rules.

 We now describe our high-level proof strategy and defer the full proof of Theorem \ref{thm:01real} to Appendix \ref{appdx:proof_realizable}. Our lower bound involves picking the worst-case shattered tree with the largest branching factor and traversing down this tree uniformly at random to generate the sequence of states. For such a sequence, the learner can do no better than random guessing at nodes where the branching occurred, yielding the lower bound $\operatorname{C}_T(\Fcal)/2$.  Next, for our upper bound, we first define a localized complexity measure $\operatorname{C}_T(\Fcal, x_0)$, where we only consider shattered trees rooted at the revealed initial state $x_0$. Our minimax learner is then a version space algorithm that predicts the state that will result in the largest reduction in the complexity measure if a mistake occurs. This learner is a generalization of the celebrated Standard Optimal Algorithm due to \cite{Littlestone1987LearningQW}.

\subsection{Minimax Rates in the Realizable Setting}
While Theorem \ref{thm:01real} provides a quantitative and qualitative characterization of realizable learnability, it does not shed light on how $\inf_{\mathcal{A}}\operatorname{M}_{\mathcal{A}}(\mathcal{F}, T)$ may depend on the time horizon $T$. In online classification with the 0-1 loss, the seminal work by \cite{Littlestone1987LearningQW} and \cite{DanielyERMprinciple} show that only two rates are possible: $\Theta(T)$ and $\Theta(1)$. That is, if a hypothesis class is online learnable in the realizable setting, then it is learnable with a constant mistake bound (i.e. the Littlestone dimension). Perhaps surprisingly, this is not the case for learning dynamical systems in a strong sense: \textit{every} rate is possible. 

\begin{theorem}\label{thm:everyrate}
\noindent For every $S \subset \mathbb{N} \cup \{0\}$, there exists $\mathcal{F}_S \subseteq \mathcal{\mathbbm{Z}}^{\mathbbm{Z}}$ such that $\operatorname{C}_{T}(\mathcal{F}_S) = \sup_{n \in \mathbb{N} \cup \{0\}} |S \cap \{n, n+1, ..., n+T-1\}|.$
\end{theorem}

Theorem \ref{thm:everyrate}, proved in Appendix \ref{app:everyrate}, along with Theorem \ref{thm:01real} implies that any minimax rate in the realizable setting is possible. As an example, suppose we would like to achieve the rate $\Theta(T^{\alpha})$ for some $\alpha < 1$. Then, picking $S = \{\lfloor  t^{\frac{1}{\alpha}} \rfloor : t \in \mathbb{N} \cup \{0\}\}$ suffices since $\sup_{x \in \mathbb{N} \cup \{0\}} |S \cap \{n, n+1, ..., n+T-1\}| = |\{\lfloor  t^{\frac{1}{\alpha}} \rfloor : t \in \mathbb{N} \cup \{0\}\} \cap \{0, 1, ..., T-1\}| = \Theta(T^{\alpha})$. Likewise, one can get logarithmic rates by picking $S = \{2^t : t \in \mathbb{N} \cup \{0\}\}$ and constant rates by picking $S \subset \mathbb{N} \cup \{0\}$ such that $|S| < \infty$. In light of Theorem \ref{thm:everyrate} and the fact that only constant mistake bounds are possible for online multiclass classification, it is natural to ask \textit{when} one can achieve constant mistake bounds for learning dynamical systems. To answer this question, we introduce a new combinatorial dimension termed the Branching dimension.

\begin{definition} [Branching dimension] \label{def:dim}
\noindent The \emph{Branching dimension}, denoted $\operatorname{Bd}(\mathcal{F})$, is the smallest natural number $d \in \mathbb{N}$ such that for every shattered trajectory tree $\mathcal{T}$, we have $\operatorname{B}(\mathcal{T}) \leq d$. If for every $d \in \mathbb{N}$, there exists a shattered trajectory tree $\mathcal{T}$ with $\operatorname{B}(\mathcal{T}) > d$, we say $\operatorname{Bd}(\mathcal{F}) = \infty$. 
\end{definition}

Theorem \ref{thm:minimaxconst}, proved via non-constructive arguments in Appendix \ref{app:minimaxconst}, shows that $\inf_{\mathcal{A}}\operatorname{M}_{\mathcal{A}}(T, \mathcal{F}) = \Theta(1)$ if and and only if $\operatorname{Bd}(\mathcal{F}) < \infty$. 

\begin{theorem} [Constant Minimax Expected Mistakes] \label{thm:minimaxconst}
\noindent For any $\mathcal{F} \subseteq \mathcal{X}^{\mathcal{X}}$, we have (i) $\inf_{\mathcal{A}}\operatorname{M}_{\mathcal{A}}(T, \mathcal{F}) \leq \operatorname{Bd}(\mathcal{F})$ and (ii) if $\operatorname{Bd}(\mathcal{F}) = \infty$, then $\inf_{\mathcal{A}}\operatorname{M}_{\mathcal{A}}(T, \mathcal{F}) = \omega(1)\footnote{Recall that $f(n) = \omega(1)$ if for all $k > 0$, there exists $n_k$, such that for all $n > n_k$ we have $f(n) > k$.}.$

\end{theorem}

\subsection{Relations to PAC and Online Multiclass Classification} \label{sec:relations}

By studying abstract state spaces $\Xcal$ and evolutions function classes $\Fcal \subseteq \Xcal^{\Xcal}$, we can also compare realizable learnability of dynamical systems to existing notions of realizable learnability in the well-known PAC and online classification settings. Our main result relates the evolution complexity to the DS and Littlestone dimensions, which characterize PAC and online classification respecitively. 

\begin{theorem}[Relations to the DS and Littlestone dimension] \label{thm:dimrel} The following statements are true.
\begin{itemize}
\item[(i)] There exists  $\mathcal{F} \subseteq \mathcal{X}^{\mathcal{X}}$ such that $\operatorname{DS}(\mathcal{F}) = \infty$ but $\operatorname{C}_T(\mathcal{F}) = \Theta(\log(T)).$
\item[(ii)] There exists  $\mathcal{F} \subseteq \mathcal{X}^{\mathcal{X}}$ such that $\operatorname{DS}(\mathcal{F}) = 1$ but $\operatorname{C}_T(\mathcal{F}) = T$.
\item[(iii)] For any $\mathcal{F} \subseteq \mathcal{X}^{\mathcal{X}}$, we have that $\operatorname{C}_T(\mathcal{F}) \leq \operatorname{L}(\mathcal{F})$.
\item[(iv)] There exists $\mathcal{F}  \subseteq \mathcal{X}^{\mathcal{X}}$ such that $\operatorname{L}(\mathcal{F}) = \infty$ but $\operatorname{C}_T(\mathcal{F}) = 1$. 
\end{itemize}
\end{theorem}

 The proof of Theorem \ref{thm:dimrel} is in Appendix \ref{app:relations}. Parts (i) and (ii) show that the finiteness of the DSdim is neither necessary nor sufficient for learning dynamical systems in the realizable setting. On the other hand, parts (iii) and (iv) show that finite Ldim is sufficient but not necessary for learning dynamical systems in the realizable setting. Overall, learning dynamical systems is always easier than online multiclass classification, but can be both easier and harder than multiclass PAC classification. The proof of (i), (ii), and (iv) are combinatorial in nature, while the proof of (iii) involves reducing learning-to-predict in dynamical systems to online multiclass classification. 

 \subsection{Examples}
 In this section, we establish the minimax rates for discrete linear systems and linear Boolean networks. The proof of Theorem \ref{thm:linsys} is in Appendix \ref{app:linearsystems}.

 \begin{theorem}[Linear Systems] \label{thm:linsys}
 \begin{itemize} 
 \item[(i)] Let $\Xcal = \mathbb{Z}^{n}$ and $r < n$.  For $\Fcal = \{x \mapsto Wx\, :\, W \in \integers^{n \times n}, \emph{\text{ rank}}(W) \leq r\} $ and $T > r$, we have $ \operatorname{C}_{T}(\Fcal) = r+1.$

 \item[(ii)] Let $\Xcal = \{0, 1\}^{n}$ and $T \geq n$.  For $\Fcal = \{x \mapsto Wx\, (\emph{\text{mod }} 2)\, :\, W \in \integers^{n \times n}\} $, we have $ \operatorname{C}_{T}(\Fcal) = n.$ 
 \item[(iii)] Let $\Xcal = \{0, 1\}^{n}$ and $T \geq n$. For  $\Fcal = \{x \mapsto \indicator\{Wx > 0\}\, :\, W \in \{0, 1\}^{n \times n}\}$,  we have $n \leq \operatorname{C}_{T}(\Fcal) \leq n^2$.
 \end{itemize}
 \end{theorem}

Thresholded Boolean networks have been used to model genetic regulatory dynamics \citep{mendoza1998dynamics} and  social networks \citep{kempe2003maximizing}. Modulo Boolean networks have been studied by \cite{chandrasekhar2023stability} in the context of stability. 


\section{Agnostic Learnability}
\subsection{Markovian Regret} \label{sec:agnostic}
In this section, we go beyond the realizable setting and consider the case where nature may reveal a trajectory that is not consistent with any evolution function in the class. 
Our main result in this section establishes bounds on the minimax expected Markovian regret.

\begin{theorem} [Minimax Expected Markovian Regret] \label{thm:01agn}
\noindent For any $\mathcal{F} \subseteq \mathcal{X}^{\mathcal{X}}$, 

$$ \max\left\{\frac{\operatorname{L}(\mathcal{F})}{18}, \frac{\sqrt{T}}{16\sqrt{3}}\right\}\leq \inf_{\mathcal{A}}\, \operatorname{MR}_{\mathcal{A}}(T, \mathcal{F}) \leq  \operatorname{L}(\mathcal{F})+ \sqrt{T\, \operatorname{L}(\mathcal{F}) \log{T}}.$$
\end{theorem}

\noindent Theorem \ref{thm:01agn} shows that the finiteness of the Littlestone dimension of $\Fcal$ is both necessary and sufficient for agnostic learnability.  This is in contrast to Theorem \ref{thm:dimrel}, which shows that the finiteness of the Littlestone dimension of $\Fcal$ is sufficient but \emph{not necessary} for realizable learnability. Thus, Theorem \ref{thm:01agn} and \ref{thm:dimrel} imply that realizable and agnostic learnability are not equivalent. 

\begin{corollary} [Realizable Learnability $\not\equiv$ Agnostic Learnability under Markovian Regret]
\noindent There exists a class $\mathcal{F} \subseteq \mathcal{X}^{\mathcal{X}}$ such that $\mathcal{F}$ is learnable in the realizable setting but not in the agnostic setting under Markovian regret. 
\end{corollary}
One such class exhibiting the separation is the thresholds $\Fcal = \{x \mapsto \indicator\{x \geq a\} \, :\, a \in (0,1)\}$ used in the proof of part (iv) in Theorem \ref{thm:dimrel}.  Beyond the qualitative separation of realizable and agnostic learnability under Markovian regret, we also observe a quantitative separation in terms of possible minimax rates. Recall that Theorem \ref{thm:everyrate} shows that every rate is possible in the realizable setting. However, Theorem \ref{thm:01agn} shows that only two types of rates are possible in the agnostic setting: $\Theta(T)$ whenever $\operatorname{L}(\Fcal)=\infty$ and $\tilde{\Theta}(\sqrt{T})$  whenever $\operatorname{L}(\Fcal)< \infty$. More precisely, when $\sqrt{T} \geq  \operatorname{L}(\Fcal)$, we have a lower bound of $\Omega(\sqrt{T})$
and an upper bound of $O(\sqrt{T \operatorname{L}(\Fcal) \log{T}})$. This raises the natural question of what the right minimax rate is. In Appendix \ref{app:tight_up_thm01agn}, we provide an evolution class and establish a lower bound of $\Omega(\sqrt{T \operatorname{L}(\Fcal)})$, showing that the upper bound is tight up to $\sqrt{\log{T}}$. 

 Our proof of the upper bound in Theorem \ref{thm:01agn} reduces learning dynamical systems to online multiclass classification and uses a result due to \cite{hanneke2023multiclass}. To prove the lower bound $\frac{\operatorname{L}(\mathcal{F})}{18}$, we construct a hard stream by carefully sampling a random path down a Littlestone tree of depth $\operatorname{L}(\mathcal{F})$. To prove the lower bound of $\frac{\sqrt{T}}{16\sqrt{3}}$, we construct a hard randomized stream using just two different evolution functions in $\mathcal{F}$. The full proof is in Appendix \ref{app:thm01agn}. 

\subsection{Flow Regret}

The necessity of the Littlestone dimension for agnostic learnability under Markovian regret is quite restrictive. For example, a simple (but unnatural) class like one-dimensional thresholds $\mathcal{F} = \{x \mapsto \mathbbm{1}\{x \leq a\}: a \in (0, 1)\}$ has $\operatorname{L}(\mathcal{F}) = \infty$ but $\operatorname{C}_T(\mathcal{F}) = 1$. The key idea in the lower bound of Theorem \ref{thm:02agn} is that the adversary can simulate the online multiclass classification game, where finiteness of the Littlestone dimension is necessary, by ``giving up" every other round. This is possible because of the definition of Markovian regret. In particular, the evaluation of the ``best-fixed evolution function in hindsight"  under Markovian regret is only penalized on one-step prediction error but not on long-term consistency of the generated dynamics starting from the initial state $x_0$.

This motivates the following natural question. Which evolution classes $\mathcal{F} \subseteq \mathcal{X}^{\mathcal{X}}$ are agnostic learnable under Flow regret? Is the finiteness of Ldim still necessary? Theorem \ref{thm:02agn}, whose proof can be found in Appendix \ref{app:prfthmagn}, provides partial answers to these question by bounding the minimax expected Flow regret for classes where $\sup_{x \in \mathcal{X}} |\mathcal{F}(x)|$ is uniformly bounded.  

\begin{theorem} [Minimax Expected Flow Regret] \label{thm:02agn} For any ordered set $\Xcal$ and $\mathcal{F} \subseteq \mathcal{X}^{\mathcal{X}}$, 

$$\frac{\operatorname{C}_T(\mathcal{F})}{2} \leq \inf_{\mathcal{A}}\operatorname{FR}_{\mathcal{A}}(T, \mathcal{F}) \leq \operatorname{C}_T(\mathcal{F}) + \sqrt{\operatorname{C}_T(\mathcal{F}) \, T \ln\Bigl(\frac  { T\, K_{\Fcal}  }{\operatorname{C}_T(\mathcal{F})}\Bigl)}.$$

\noindent where $ K_{\Fcal} = \sup_{x \in \mathcal{X}} |\mathcal{F}(x)|$. Moreover, both the lower- and upper bound can be tight. 
\end{theorem}

Note that the upper bound becomes vacuous when $\sup_{x \in \mathcal{X}} |\mathcal{F}(x)| = \infty$, and finding a general characterization of Flow regret learnability remains an open question. Unlike agnostic learnability under Markovian regret, the finiteness of the Ldim is not necessary for agnostic learnability under Flow regret. Indeed, while the class of one-dimensional thresholds is not agnostic learnable under Markovian regret, it is agnostic learnable under Flow regret. Although Markovian learnability is not necessary for Flow regret learnability, when $K_{\Fcal}<\infty$, learnability under Markovian regret is sufficient learnability under flow regret. To see this, recall that Theorem \ref{thm:01agn} states that learnability under Markovian regret implies $\operatorname{L}(\Fcal)<\infty$. Then, since part (iii) of Theorem \ref{thm:dimrel} states $\operatorname{C}_T(\Fcal)\leq \operatorname{L}(\Fcal)$, we can use Theorem \ref{thm:02agn} to infer that $\Fcal$ is also learnable under flow regret with regret $\leq \operatorname{L}(\Fcal) + \sqrt{\operatorname{L}(\Fcal) T \ln(T K_{\Fcal})}$.

 Theorem \ref{thm:02agn} shows that realizable and agnostic learnability under Flow regret are equivalent as long as $\sup_{x \in \mathcal{X}} |\mathcal{F}(x)| < \infty$.  But this equivalence breaks down when the projection sizes are unbounded, as shown by Theorem \ref{thm:sepflowregret}.

 \begin{theorem} [Realizable learnability $\not\equiv$ Agnostic Learnability under Flow Regret] \label{thm:sepflowregret}
 \noindent There exists an ordered set $\Xcal$ and $\mathcal{F}\subseteq \mathcal{X}^{\mathcal{X}}$ such that (i) $\inf_{\Acal}\operatorname{M}_{\Acal}(T, \mathcal{F}) \leq 3$ but (ii) $ \inf_{\Acal} \operatorname{FR}_{\Acal}(T, \mathcal{F})  \geq \frac{T}{6}$.
 \end{theorem}

To prove Theorem \ref{thm:sepflowregret} (see Appendix \ref{app:proofofsep}), we construct a class $\mathcal{F} \subseteq \mathcal{X}^{\mathcal{X}}$ such that on large subset of states in $\mathcal{X}^{\prime} \subset \mathcal{X}$, every function $f \in \mathcal{F}$ effectively reveals its identity. \\

\section{Discussion and Future directions}
In this work, we studied the problem of learning-to-predict in discrete-time dynamical systems under the $0$-$1$ loss. A natural extension is to consider continuous state spaces with real-valued losses. For example, one can take $\Xcal$ to be a bounded subset of a Hilbert space and consider the squared norm as the loss function. Another natural extension is to consider learnability under partial observability, where the learner only observes some transformation $\phi(x_t)$ instead of the true state $x_t$. Such feedback model is standard in prediction for linear dynamical systems \citep{hazan2018spectral}. It is also natural to study the learnability of function classes where the output of the evolution rules $f: \Xcal^p \to \Xcal$, depend on the previous $p > 1$ states (e.g. the $p$-th order VAR model).  Lastly, the learning algorithms in this work are \textit{improper}: they use evolution functions that may not lie in $\mathcal{F}$ to make predictions.  This might be undesirable as improper learning algorithms may be incompatible with downstream system identification and control tasks. To this end, characterizing \emph{proper} learnability of dynamical systems is an important future direction.

\acks{AT and US acknowledge the support of NSF via grant DMS-2413089. US also acknowledges the support of Rackham International Student Fellowship. VR acknowledges the support of the NSF GRFP. We thank Chinmaya Kaushik for pointing out a technical result, which helped us to prove part (i) Theorem \ref{thm:linsys}.}

\bibliography{references}

\newpage

\appendix

\section{Proofs for Realizable learnability}\label{appdx:proof_realizable}

\begin{proof} (of lower bound of Theorem \ref{thm:01real}). Let $\mathcal{F} \subseteq \mathcal{X}^{\mathcal{X}}$ be any evolution class and $T \in \mathbb{N}$ be the time horizon. Let $\mathcal{A}$ be any randomized learner. Our goal will be to construct a hard realizable trajectory $\{x_t\}_{t=0}^T$ such that $\mathcal{A}$'s expected number of mistakes is at least $\frac{\operatorname{C}_T(\mathcal{F})}{2}$. Without loss of generality, suppose $d:= \operatorname{C}_T(\mathcal{F}) > 0$ as otherwise the lower bound holds trivially. Then, by definition of the Evolution complexity, there exists a trajectory tree $\mathcal{T}$ of depth $T$ shattered by $\mathcal{F}$ with branching factor at least $d$. This means that for every path $\sigma \in \{-1, 1\}^T$ down $\mathcal{T}$, we have $\sum_{t=1}^{T} \mathbbm{1}\{\mathcal{T}_t((\sigma_{<t}, -1)) \neq  \mathcal{T}_t((\sigma_{<t}, +1))\} \geq d$. 

Let $\sigma \sim \{-1, 1\}^T$ denote a random path down $\mathcal{T}$ and consider the trajectory $\mathcal{T}_0 \cup \{\mathcal{T}_t(\sigma_{\leq t})\}_{t=1}^T$. Define $\Tcal_{<t}(\sigma_{<t})  := (\Tcal_0, \Tcal_{1}(\sigma_1), \ldots, \Tcal_{t-1}(\sigma_{<t}))$. Then, observe that 

\begin{align*}
\mathbbm{E}\left[\sum_{t=1}^T \mathbbm{1}\{\mathcal{A}(\mathcal{T}_{<t}(\sigma_{<t}))\neq \mathcal{T}_t(\sigma_{\leq t})\} \right] &= \sum_{t=1}^T \mathbbm{E}\Big[\mathbbm{E}\left[\mathbbm{1}\left\{\mathcal{A}\big(\mathcal{T}_{<t}(\sigma_{<t})\big) \neq \mathcal{T}_t\big((\sigma_{<t}, \sigma_t)\big) \right\} \, \Big|\, \sigma_{<t} \right]\Big]\\
&\geq \frac{1}{2}\sum_{t=1}^T \mathbbm{E}\left[ \mathbbm{E}\Big[\mathbbm{1}\left\{ \mathcal{T}_t\big((\sigma_{<t}, -1)\big) \neq \mathcal{T}_t\big((\sigma_{<t}, +1)\big) \right\} \,\Big|\, \sigma_{<t} \Big] \right]\\
&= \frac{1}{2}\, \mathbbm{E}\left[\sum_{t=1}^T \mathbbm{1} \left\{ \mathcal{T}_t((\sigma_{<t}, -1)) \neq \mathcal{T}_t\big((\sigma_{<t}, +1)\big) \right\} \right] \geq \frac{d}{2}
\end{align*}

where the first inequality follows from the fact that $\sigma_t \sim \{-1, 1\}$. \end{proof}

\begin{proof} (of upper bound of Theorem \ref{thm:01real}). Let $\mathcal{F} \subseteq \mathcal{X}^{\mathcal{X}}$ be any evolution function class and let $T \in \mathbb{N}$ be the time horizon. Our goal will be to construct a deterministic learner $\mathcal{A}$ such that for any realizable trajectory $\{x_t\}_{t=0}^T$, $\mathcal{A}$ makes at most $\operatorname{C}_T(\mathcal{F})$ mistakes. To that end, it will be useful to define an instance-dependent version of the evolution complexity. 

Given a state $x \in \mathcal{X}$, we say $\mathcal{T}$ is a trajectory tree rooted at $x$ if the root node is labeled by $x$. For an initial state $x_0 \in \mathcal{X}$ and evolution class $V \subseteq \mathcal{F}$, let 
$$\operatorname{C}_T(V, x_0) := \sup\{\operatorname{B}(\Tcal) \, \mid \, \Tcal \in \Scal(V, T) \text{ and } \Tcal \text{ rooted at } x_0 \}$$
\noindent denote an instance-dependent Evolution complexity of $V$, where $\Scal(V, T)$ is set of all trajectory trees of depth $T$ shattered by $V$.  Note that $\operatorname{C}_T(V) = \sup_{x_0 \in \mathcal{X}} \operatorname{C}_T(V,x_0)$. $\mathcal{A}$ is a version-space algorithm that uses the instance-dependent Evolution complexity to make its prediction such that whenever $\mathcal{A}$ errs, the instance-dependent Evolution complexity decreases. In this way, $\operatorname{C}_T(V, x_0)$ acts as a potential function. Algorithm \ref{alg:real} formalizes this idea. 

\begin{algorithm}[H]
\setcounter{AlgoLine}{0}
\caption{Deterministic Realizable Algorithm}\label{alg:real}
\KwIn{Initial state $x_0 \in \mathcal{X}$}

Let $V_{1} = \mathcal{F}$

\For{$t = 1,...,T$} {

    For $x \in \mathcal{X}$, define $V_t^x = \{f \in V_t: f(x_{t-1}) = x\}.$

    \uIf{$\{f(x_{t-1}): f \in V_t\} = \{x\}$}{
        Predict $\hat{x}_t = x$.
    }

    \uElse{

     Predict $\hat{x}_t \in \argmax_{x \in \mathcal{X}}\operatorname{C}_{T-t}(V_{t}^x,  x).$
    }

    Receive $x_t$ and update $V_{t+1} \leftarrow V_t^{x_t}.$ 
}
\end{algorithm}

We now show that Algorithm \ref{alg:real} makes at most $\operatorname{C}_T(\mathcal{F})$ mistakes on any realizable trajectory. Let $\{x_t\}_{t=0}^T$ denote the realizable trajectory to be observed. It suffices to show that 

\begin{equation} \label{eq:real}
    \operatorname{C}_{T-t}(V_{t+1}, x_{t}) \leq \operatorname{C}_{T-t+1}(V_{t}, x_{t-1}) - \mathbbm{1}\{x_t \neq \hat{x}_t\}
\end{equation}

for all $t \in [T]$. To see why this is true, note that (\ref{eq:real}) implies  

$$\sum_{t=1}^T \operatorname{C}_{T-t}(V_{t+1}, x_{t}) \leq \sum_{t=1}^T\operatorname{C}_{T-t+1}(V_{t}, x_{t-1}) - \sum_{t=1}^T \mathbbm{1}\{x_t \neq \hat{x}_t\}.$$

Rearranging, we get that 

\begin{align*}
    \sum_{t=1}^T \mathbbm{1}\{x_t \neq \hat{x}_t\} &\leq \sum_{t=1}^T\operatorname{C}_{T-t+1}(V_{t},x_{t-1}) - 
    \sum_{t=1}^T \operatorname{C}_{T-t}(V_{t+1}, x_{t})\\
    &= \sum_{t=1}^T \Bigl(\operatorname{C}_{T-t+1}(V_{t}, x_{t-1}) - \operatorname{C}_{T-t}(V_{t+1}, x_{t})\Bigl)\\
    &= \operatorname{C}_T(V_{1}, x_{0}) - \operatorname{C}_0(V_{T+1},x_{T}) \leq \operatorname{C}_T(\mathcal{F}).
\end{align*}

as needed. To prove (\ref{eq:real}), we need to consider three cases: (a) $\operatorname{C}_{T-t+1}(V_{t}, x_{t-1}) > 0$ and $\hat{x}_t \neq x_t$, (b) $\operatorname{C}_{T-t+1}(V_{t}, x_{t-1}) > 0$ and $\hat{x}_t = x_t$, and (c) $\operatorname{C}_{T-t+1}(V_{t}, x_{t-1}) = 0$.  

Starting with (a), let $t \in [T]$ such that  $\operatorname{C}_{T-t+1}(V_{t}, x_{t-1}) > 0$ and $\hat{x}_t \neq x_t$. We need to show that $\operatorname{C}_{ T-t}(V_{t+1}, x_t) < \operatorname{C}_{T-t+1}(V_{t}, x_{t-1})$. Suppose for the sake of contradiction that $\operatorname{C}_{T-t}(V_{t+1}, x_{t}) \geq \operatorname{C}_{T-t+1}(V_{t}, x_{t-1}) := d$. Then, by the prediction rule, we have that $\operatorname{C}_{T-t}(V^{\hat{x}_t}_{t}, \hat{x}_t) \geq \operatorname{C}_{T-t}(V^{x_t}_{t}, x_t) \geq d$. By definition of the instance-dependent Evolution complexity, we are guaranteed the existence of a trajectory tree $\mathcal{T}_{x_t}$ of depth $T-t$, rooted at $x_t$, shattered by $V_t^{x_t}$ with branching factor at least $d$ and a tree $\mathcal{T}_{\hat{x}_t}$ of depth $T-t$, rooted at $\hat{x}_t$, shattered by $V_t^{\hat{x}_t}$ with branching factor at least $d$. Consider a binary tree $\mathcal{T}$ whose root node $v$ is labeled by $x_{t-1}$ and left and right subtrees are $\mathcal{T}_{x_t}$ and $\mathcal{T}_{\hat{x}_t}$ respectively. Then, observe that $\mathcal{T}$ is a trajectory tree of depth $T-t+1$, rooted at $x_{t-1}$, shattered by $V_t$ with branching factor at least $d+1$ (because $x_t \neq \hat{x}_t$). This contradicts our assumption that $\operatorname{C}_{T-t+1}(V_{t}, x_{t-1}) = d$. Thus, we must have $\operatorname{C}_{ T-t}(V_{t+1}, x_t) < \operatorname{C}_{T-t+1}(V_{t}, x_{t-1})$.

Moving to (b), let $t \in [T]$ be such that $\hat{x}_t = x_t$. Then, we need to show that  $d: = \operatorname{C}_{T-t}(V_{t+1}, x_t) \leq \operatorname{C}_{T-t+1}(V_{t}, x_{t-1}).$ If $d = 0$, the inequality holds trivially. Thus, assume $d > 0$. Then, by definition, there exists a trajectory tree $\mathcal{T}$ of depth $T-t$, rooted at $x_t$, shattered by $V_{t+1}$ with branching factor at least $d$. Consider a binary tree $\tilde{\mathcal{T}}$ whose root node $v$ is labeled by $x_{t-1}$ and left and right subtrees are both $\mathcal{T}$. Then since $V_{t+1} = V_t^{x_t} \subseteq V_t$, we have that $\tilde{\mathcal{T}}$ is a trajectory tree of depth $T-t+1$, rooted at $x_{t-1}$, shattered by $V_{t}$  with branching factor at least $d$. Thus, we must have $\operatorname{C}_{T-t+1}(V_{t}, x_{t-1}) \geq d$.

Finally for (c), let $t \in [T]$ be such that $\operatorname{C}_{T-t+1}(V_{t}, x_{t-1}) = 0$. Using the same logic as (b), $\operatorname{C}_{T-t}(V_{t+1}, x_t) = 0$. Thus, to prove that \eqref{eq:real} holds, it suffices to show that $x_t = \hat{x}_t$.  To do so, we will show that the projection size $|\{f(x_{t-1}): f \in V_t\}| = 1$. Thus, Algorithm \ref{alg:real} does not make a mistake. Suppose for the sake of contradiction that $|\{f(x_{t-1}): f \in V_t\}| \geq 2$. Then, there exists two functions $f_1, f_2 \in V_t$ such that $f_1(x_{t-1}) \neq f_2(x_{t-1})$. Consider a binary tree $\mathcal{T}$ of depth $T-t+1$ where the root node is labeled by $x_{t-1}$, and its left and right child by $f_1(x_{t-1})$ and $f_2(x_{t-1})$ respectively. Label all remaining internal nodes in the left subtree of the root node such that every path is shattered by $f_1$. Likewise, for the right subtree of the root node, label every internal node such that every path is shattered by $f_2$. Following this procedure, $\mathcal{T}$ is a trajectory tree of depth $T-t+1$, rooted at $x_{t-1}$, shattered by $V_t$ with branching factor at least $1$ (because $f_1(x_{t-1}) \neq f_2(x_{t-1})$). Thus,  $\operatorname{C}_{T-t+1}(V_{t}, x_{t-1}) \geq 1$, which is a contradiction. \end{proof}

\section{Proof of Theorem \ref{thm:everyrate}} \label{app:everyrate}

Let $S \subset \mathbb{N} \cup \{0\}$ be an arbitrary subset of the extended natural numbers. For every $\sigma \in \{-1, 1\}^{\mathbb{N} \cup \{0\}}$, define the evolution function 

$$
f_{\sigma}(x) = \Bigl(\sigma_{|x|}\mathbbm{1}\{|x| \in S\} + \mathbbm{1}\{|x| \notin S \}\Bigl) (|x|+1)
$$

\noindent and consider the class $\mathcal{F}_S = \Bigl\{f_{\sigma}: \sigma \in \{-1, 1\}^{\mathbb{N} \cup \{0\}}\Bigl\}.$ By construction, functions in $\mathcal{F}_S$ only disagree on states in $S$ and their negation. Moreover, given any initial state $x_0 \in \mathbbm{Z}$ and a time horizon $T \in \mathbb{N}$, the trajectory of any evolution in $\mathcal{F}$ is $\epsilon_1 (|x_0| + 1), ..., \epsilon_T (|x_0| + T)$ for some $\epsilon \in \{-1, 1\}^T.$

We now show that $\operatorname{C}_T(\Fcal) \leq \sup_{x_0 \in \mathbb{Z}}|S \cap \{|x_0|, ..., |x_0| + T - 1\}|$. Let $\mathcal{T}$ be any trajectory tree of depth $T$ shattered by $\mathcal{F}$. It suffices to show that $\operatorname{B}(\mathcal{T}) \leq |S \cap \{|\Tcal_0|, ..., |\Tcal_0| + T - 1\}|$. Fix any path $\epsilon \in \{-1, 1\}^T$ down $\mathcal{T}$ and consider the sequence of states $\mathcal{T}_0, ..., \mathcal{T}_{T}(\epsilon_{\leq T})$.  Note that $\Tcal_t((\epsilon_{<t}, -1)) \neq \Tcal_t((\epsilon_{<t}, +1))$ only if $|\Tcal_{t-1}(\epsilon_{<t})| \in S$. Thus, 

\begin{align*}
\sum_{t=1}^{T} \mathbbm{1}\{\mathcal{T}_t((\epsilon_{<t}, -1)) \neq  \mathcal{T}_t\big((\epsilon_{<t}, +1)\big) \} &\leq \sum_{t=1}^T \mathbbm{1}\{|\Tcal_{t-1}(\epsilon_{<t})| \in S\} \\
&= \sum_{t=0}^{T-1} \mathbbm{1}\{|\Tcal_{t}(\epsilon_{\leq t})| \in S\}\\
&= \mathbbm{1}\{|\Tcal_0| \in S\} + \sum_{t=1}^{T-1} \mathbbm{1}\{|\Tcal_{t}(\epsilon_{\leq t})| \in S\}
\end{align*}

Moreover, since the path $\epsilon$ is shattered by $\mathcal{F}_S$, it must be the case that for every $t \in [T]$, we have that $|\mathcal{T}_t(\epsilon_{\leq t})| = |\mathcal{T}_0| + t$. Thus, 

\begin{align*}
\sum_{t=1}^{T} \mathbbm{1}\{\mathcal{T}_t((\epsilon_{<t}, -1)) \neq  \mathcal{T}_t\big((\epsilon_{<t}, +1)\big) \} &\leq \mathbbm{1}\{|\Tcal_0| \in S\} + \sum_{t=1}^{T-1} \mathbbm{1}\{|\Tcal_{t}(\epsilon_{\leq t})| \in S\}\\
&= \sum_{t=0}^{T-1} \mathbbm{1}\{|\mathcal{T}_0| + t \in S\}\\
&= |S \cap \{|\Tcal_0|, ..., |\Tcal_0| + T - 1\}|, 
\end{align*}

\noindent Taking the supremum of both sides with respect to $\mathcal{T}_0$ completes the proof of the upper bound. 

To prove the lower bound, fix $x \in \mathbb{Z}$ and consider the following trajectory tree $\Tcal^x$ of depth $T$. Let $\mathcal{T}^x_0 = x$. For every path $\epsilon \in \{-1, 1\}^T$ and $t \in [T]$, let 

$$\mathcal{T}^x_{t}(\epsilon_{\leq t}) = \begin{cases}
			\epsilon_t (|\Tcal^x_0| + t), & \text{if} |\Tcal^x_0| + t - 1 \in S.\\
            |\Tcal^x_0| + t & \text{if } |\Tcal^x_0| + t - 1 \notin S.
		 \end{cases}$$

Note $|\mathcal{T}^x_{t}(\epsilon_{\leq t})| = |\mathcal{T}^x_0| + t$ for all $t \in [T]$. Moreover, for every path $\epsilon \in \{-1, 1\}^T$, there exists a function $\sigma \in \{-1, 1\}^{\mathbbm{N} \cup \{0\}}$ such that $\sigma_{|\mathcal{T}^x_0|} = \epsilon_1$ and $\sigma_{|\mathcal{T}^x_{t}(\epsilon_{\leq t})|} = \epsilon_{t+1}$ for all $t \in [T-1].$ Thus, the function $f_{\sigma} \in \mathcal{F}_S$ shatters the path $\epsilon$ and the tree $\Tcal^x$ is shattered by $\mathcal{F}_S$. We claim that $\operatorname{B}(\mathcal{T}^x) = |S \cap \{|\Tcal^x_0|, ..., |\Tcal^x_0| + T - 1\}.$ To see this, observe that for every path $\epsilon \in \{-1, 1\}^T$, we have 

\begin{align*}
\sum_{t=1}^{T} \mathbbm{1}\{\mathcal{T}^x_t((\epsilon_{<t}, -1)) \neq  \mathcal{T}^x_t\big((\epsilon_{<t}, +1)\big) \} &= \sum_{t=1}^T \mathbbm{1}\{|\Tcal^x_0| + t - 1 \in S\}\\
&= |S \cap \{|x_0|, ..., |x_0| + T - 1\}|.
\end{align*}

Thus, $\operatorname{B}(\mathcal{T}^x) = |S \cap \{|x_0|, ..., |x_0| + T - 1\}|$ and $\operatorname{C}_T(\mathcal{F}) \geq \sup_{x \in \mathbb{Z}}\operatorname{B}(\mathcal{T}^x) = \sup_{x \in \mathbb{Z}} |S \cap \{|x|, ..., |x| + T - 1\}|.$ This completes the proof.

\section{Proof of Theorem \ref{thm:minimaxconst}}\label{app:minimaxconst}





The proof of (i) follows from Theorem \ref{thm:01real} and the fact that $\operatorname{C}_T(\mathcal{F}) \leq \operatorname{Bd}(\mathcal{F})$. To prove (ii), observe that by Theorem \ref{thm:01real}, it suffices to show that when $\operatorname{Bd}(\mathcal{F}) = \infty$, we have that $\operatorname{C}_T(\mathcal{F}) = \omega(1)$. If $\operatorname{Bd}(\mathcal{F}) = \infty$, then for every $d \in \mathbb{N}$, there exists a shattered trajectory tree $\mathcal{T}$ such that $\operatorname{B}(\mathcal{T}) > d$.  Thus, for every $d \in \mathbb{N}$, there exists a depth $d^{\prime} \in \mathbb{N}$, such that for every $T \geq d^{\prime}$, there exists a shattered tree $\mathcal{T}$ of depth $T$ with $\operatorname{B}(\mathcal{T}) > d$. In other words, for every $d \in \mathbb{N}$, there exists a $d^{\prime} \in \mathbb{N}$ such that for all $T \geq d^{\prime}$ we have that $\operatorname{C}_{T}(\mathcal{F}) > d$. By definition of $\omega(\cdot)$, this means that $\operatorname{C}_T(\mathcal{F}) = \omega(1)$ as needed. 

\section{Proof of Theorem \ref{thm:dimrel}} \label{app:relations}

\begin{proof} (of (i) in Theorem \ref{thm:dimrel}) To prove (i), pick $S = \{2^t: t \in \mathbb{N} \cup \{0\}\}$ and consider the function class $\mathcal{F}_S$ from Theorem \ref{thm:everyrate}. It is not too hard to see that $\operatorname{C}_T(\mathcal{F}) = \Theta(\log(T))$. In addition, one can verify that every finite subset of $S$ is a shattered set according to Definition \ref{def:dsdim}. Indeed, consider any finite subset $A \subset S$. For every $i \in [|A|]$, let $A_i$ denote the $i$'th element of $A$ after sorting $A$ in increasing order. Then, observe that by the construction of $\mathcal{F}_S$, for every sequence $\epsilon \in \{-1, 1\}^{|A|}$, there exists a function $f_{\epsilon} \in \mathcal{F}_S$ such that $f_{\epsilon}(A_i) = \epsilon_i (A_i + 1)$ for every $i \in [|A|].$ By letting $\mathcal{H} = \{f_{\epsilon}: \epsilon \in \{-1, 1\}^{|A|}\}$ in Definition \ref{def:dsdim}, one can verify that $|\mathcal{H}| = 2^{|A|} < \infty$, and for every $x \in A$ and $h \in \mathcal{H}$, there exists a $g \in \mathcal{H}$ such that $h(x) = -g(x)$ and $h(z) = g(z)$ for all $z \in A \setminus \{x\}$. Thus, $\mathcal{F}_S$ shatters $A \subset S$. Since finite subsets of $S$ can be arbitrary large, we have that $\operatorname{DS}(\mathcal{F}) = \infty$. \end{proof}

\begin{proof}(of (ii) in in Theorem \ref{thm:dimrel}) To prove (ii), let $\mathcal{X} = \mathbb{N} \cup \{\star\}$ and consider the following evolution function class. For every $\sigma \in \{-1, 1\}^{\mathbb{N}}$, define a sequence $a_{\sigma}: \mathbb{N} \rightarrow \mathbb{N}$ recursively such that $a_{\sigma}(1) = 1$ and $a_{\sigma}(n) = 2a_{\sigma}(n-1) + \frac{1+\sigma_{n-1}}{2}$ for $n \geq 2$. Equivalently, we can define the sequence $a_{\sigma}$ explicitly by $a_{\sigma}(1) = 1$ and  $a_{\sigma}(n) = 2^{n-1} + \sum_{i=1}^{n-1} \Bigl(\frac{1 + \sigma_i}{2}\Bigl) 2^{n-(i+1)}$ for $n \geq 2$. Let $S_n := \bigcup_{\sigma} \{a_{\sigma}(n)\}$ and note that $S_n = \{2^{n-1}, ..., 2^n - 1\}$ for every $n \geq 1$ and $S_n \cap S_r = \emptyset$ for all $n \neq r$.  We establish some important properties about these sequences. 
\begin{itemize}
\item[(1)] The sequence $a_{\sigma}$ is strictly monotonically increasing in its input, and hence invertible. Accordingly, given a sequence $a_{\sigma}$ and an element $x \in \text{im}(a_{\sigma})$, let $a^{-1}_{\sigma}(x)$ denote the index $n \in \mathbb{N}$ such that $a_{\sigma}(n) = x$.
\item[(2)] For every $\sigma_1, \sigma_2$, if $a_{\sigma_1}(n) = a_{\sigma_2}(r)$, then $n = r$ since $S_n \cap S_r = \emptyset$ for all $n \neq r$. 
\item[(3)] The value of $a_{\sigma}(n)$ depends only on the prefix $(\sigma_1, ..., \sigma_{n-1})$. Hence, two strings $\sigma_1, \sigma_2$ that share the same prefix up to and including index $d$ will have the property that $a_{\sigma_1}(n) = a_{\sigma_2}(n)$ for all $n \leq d+1$.
\item[(4)] If  $a_{\sigma_1}(d) = a_{\sigma_2}(d)$, then $a_{\sigma_1}(i) = a_{\sigma_2}(i)$ for all $i \leq d$. This follows by induction. For the base case, consider $i = d - 1$. If  $a_{\sigma_1}(i) \neq a_{\sigma_2}(i)$, then $a_{\sigma_1}(d) = 2a_{\sigma_1}(i) + \frac{1+\sigma_{1, i}}{2} \neq 2a_{\sigma_2}(i) + \frac{1+\sigma_{2, i}}{2} = a_{\sigma_2}(d)$ for any value of $\sigma_{1, i}$ and $\sigma_{2, i}$. For the induction step, suppose $a_{\sigma_1}(i) = a_{\sigma_2}(i)$ for some $2 \leq i < d$. Then, if $a_{\sigma_1}(i-1) \neq a_{\sigma_2}(i-1)$, we have that $a_{\sigma_1}(i) = 2a_{\sigma_1}(i-1) + \frac{1+\sigma_{1, i-1}}{2} \neq 2a_{\sigma_2}(i-1) + \frac{1+\sigma_{2, i-1}}{2} = a_{\sigma_2}(i)$ for any value of  $\sigma_{1, i-1}$ and $\sigma_{2, i-1}$. 
\end{itemize}

We now construct a function class. For every $\sigma \in \{-1, 1\}^{\mathbb{N}}$, define the evolution function 

$$f_{\sigma}(x) = \star \mathbbm{1}\{x \notin \text{im}(a_{\sigma})\} + a_{\sigma}\Bigl(a^{-1}_{\sigma}(x) + 1\Bigl)\mathbbm{1}\{x \in \text{im}(a_{\sigma})\}.$$

\noindent At a high-level, the evolution function $f_{\sigma}$ maps every state in $\mathcal{X}\setminus \text{im}(a_{\sigma})$ to $\star$ and every state in $\text{im}(a_{\sigma})$ to the next element in the sequence corresponding to $a_{\sigma}$. 

Consider the function class $\mathcal{F} = \{f_{\sigma}: \sigma \in \{-1, 1\}^{\mathbb{N}}\}.$ We now claim that $\operatorname{C}_{T}(\mathcal{F}) = T$. To see this, fix a depth $d \in \mathbb{N}$, and consider the following trajectory tree $\mathcal{T}$ of depth $d$. Let the root node be labeled by $1$, that is $\mathcal{T}_0 = 1$. For all $t \in [d]$ and $\epsilon \in \{-1, 1\}^t$, let $\mathcal{T}_t (\epsilon) = a_{\tilde{\epsilon}}(t+1)$ where $\tilde{\epsilon}$ denotes an arbitrary extension of $\epsilon$ over $\mathbb{N}$. Note that the completion can be arbitrary because the value of $a_{\sigma}(t+1)$ for any $\sigma \in \{-1, 1\}^{\naturals}$ depends only on the prefix $(\sigma_1, ..., \sigma_{t})$. One can verify that such a tree $\mathcal{T}$ is a complete binary tree of depth $d$ where the root node is labeled with $1$ and the internal nodes on depth $i \geq 1$ are labeled from left to right by $2^i, 2^i+1, ..., 2^{i+1} - 1$. Thus, it is clear that $\operatorname{B}(\mathcal{T}) = d$ since for every internal node including the root, its two children are labeled by differing states.  In addition, observe that for every path $\epsilon \in \{-1, 1\}^d$ down $\mathcal{T}$, the function $f_{\tilde{\epsilon}} \in \mathcal{F}$ shatters the associated sequence of states, where $\tilde{\epsilon}$ again is an arbitrary completion of $\epsilon$ over $\mathbb{N}$. Indeed, fix a $\epsilon \in \{-1, 1\}^d$, a completion $\tilde{\epsilon} \in \{-1, 1\}^{\mathbb{N}}$, and consider any $t \in [d]$. Then, by definition of $\mathcal{T}$, we have that $\mathcal{T}_{t-1}(\epsilon_{<t}) = a_{\tilde{\epsilon}}(t)$ and $\mathcal{T}_t(\epsilon_{\leq t}) = a_{\tilde{\epsilon}}(t+1)$. Consider the function $f_{\tilde{\epsilon}} \in \mathcal{F}$. By definition of $f_{\tilde{\epsilon}}$, we have that $f_{\tilde{\epsilon}}(a_{\tilde{\epsilon}}(t)) = a_{\tilde{\epsilon}}(t+1)$, which implies that $f_{\tilde{\epsilon}}(\mathcal{T}_{t-1}(\epsilon_{<t})) = \mathcal{T}_t(\epsilon_{\leq t})$. Since $t \in [d]$ was arbitrary, this is true for every $t \in [d]$, and thus $f_{\tilde{\epsilon}}$ shatters the path $\epsilon$ down $\mathcal{T}$ as claimed. Since $\epsilon \in \{-1, 1\}^d$ was also arbitrary, we have that the entire tree $\mathcal{T}$ is shattered by $\mathcal{F}$. Finally, since $d \in \mathbb{N}$ was arbitrary, this is true for arbitrarily large depths. Thus, $\operatorname{C}_T(\Fcal) = T$. 

We now show that $\operatorname{DS}(\mathcal{F}) = 1$ by proving that $\mathcal{F}$ cannot DS-shatter any two instances $x_1, x_2 \in \mathcal{X}$. Our proof will be in cases. First, observe that if either $x_1 = \star$ or $x_2 = \star$, then $(x_1, x_2)$ cannot be shattered since all functions in $\mathcal{F}$ will output $\star$ on either $x_1$ or $x_2$. Thus, without loss of generality, suppose both $x_1, x_2 \in \mathbb{N}$ and $x_1 < x_2$. Consider any finite subset $\mathcal{H} \subset \mathcal{F}$ and suppose there exists a function $h \in \mathcal{H}$ such that $h(x_2) \neq \star$. Then, in order to shatter $(x_1, x_2)$, there must exist a function $g \in \mathcal{H}$ such that $h(x_1) \neq g(x_1)$ but $h(x_2) = g(x_2)$.  However, if $h(x_2) = g(x_2) \neq \star$, then it must be the case that $h(x_1) = g(x_1)$. To see why, fix an instance $x \in \mathbb{N}$, and suppose $f_{\sigma_1}(x) = f_{\sigma_2}(x) \neq \star$. Then, by properties (2) and (4) above, it must be the case that $a^{-1}_{\sigma_1}(x) = a^{-1}_{\sigma_2}(x) = c$ and $a_{\sigma_1}(i) = a_{\sigma_2}(i)$ for all $i \leq c$. 
Thus, if $x_1 < x_2$ and $f_{\sigma_1}(x_2) = f_{\sigma_2}(x_2) \neq \star$, we must have that $f_{\sigma_1}(x_1) = f_{\sigma_2}(x_1)$ because either $x_1 \notin \text{im}(a_{\sigma_1}) \cup \text{im}(a_{\sigma_2}) $ or $a^{-1}_{\sigma_1}(x_1) < a^{-1}_{\sigma_1}(x_2) = a^{-1}_{\sigma_2}(x_2)$. Thus, if $\mathcal{H}$ were to satisfy the property in Definition \ref{def:dsdim}, there cannot exist a hypothesis $h \in \mathcal{H}$ such that $h(x_2) \neq \star$. However, if for every $h \in \mathcal{H}$, we have that $h(x_2) = \star$, then for every $h \in \mathcal{H}$, there cannot exist a $g \in \mathcal{H}$ such that $h(x_1) = g(x_1)$ and $h(x_2) \neq g(x_2)$.  Therefore, the two points $(x_1, x_2)$ cannot be shattered. Since $(x_1, x_2)$ and $\mathcal{H} \subset \mathcal{F}$ were arbitrary, this is true for all such points, implying that $\operatorname{DS}(\mathcal{F}) \leq 1$. Since $|\mathcal{F}| \geq 2$, we have also that $\operatorname{DS}(\mathcal{F}) \geq 1$, completing the proof that $\operatorname{DS}(\mathcal{F}) = 1$. 
\end{proof}

\begin{proof} (of (iii) in Theorem \ref{thm:dimrel})
To prove part (iii) of Theorem \ref{thm:dimrel}, we reduce learning dynamical systems to online multiclass classification.  Let $\mathcal{F} \subseteq \mathcal{X}^{\mathcal{X}}$ and $\mathcal{B}$ be any (potentially randomized) online learner for $\mathcal{F}$ for online multiclass classification with expected regret bound $R$. We will construct a learner $\mathcal{A}$ that uses $\mathcal{B}$ as a subroutine such that $\operatorname{M}_{\mathcal{A}}(T, \mathcal{F}) \leq R$.

To that end, let $(x_0, x_1, ..., x_T)$ be the \textit{realizable} stream to be observed by the learner and consider the following learning algorithm $\mathcal{A}$ which makes the same predictions as $\mathcal{B}$ while simulating the stream of labeled instance $(x_0, x_1), ..., (x_{T-1}, x_T)$ to $\mathcal{B}$. 


\begin{algorithm}[H]
\label{alg:det_SOA}
\setcounter{AlgoLine}{0}
\caption{Learning algorithm $\mathcal{A}$.}
\KwIn{Online multiclass learner $\mathcal{B}$, initial state $x_0$.}

\For{$t = 1,...,T$} {

     Pass $x_{t-1}$ to $\mathcal{B}$ and receive prediction $\hat{z}_t.$

     Predict $\hat{x}_t = \hat{z}_t$.

     Receive next state $x_t$ and update $\mathcal{B}$ using labeled instance $(x_{t-1}, x_t).$
}
\end{algorithm}



Then, observe that

\begin{equation}\label{reduction}
\mathbbm{E}\left[ \sum_{t=1}^T \mathbbm{1}\{\hat{x_t} \neq x_t\} \right] = \mathbbm{E}\left[ \sum_{t=1}^T \mathbbm{1}\{\hat{z_t} \neq x_t\}\right] \stackrel{\text{(i)}}{\leq} \inf_{f \in \mathcal{F}} \sum_{t=1}^T \mathbbm{1}\{f(x_{t-1}) \neq x_t\} + R \stackrel{\text{(ii)}}{=} R
\end{equation}

where (i) follows from the expected regret guarantee of $\mathcal{B}$ and (ii) follows from the fact that the stream of states is realizable. Thus, $\inf_{\mathcal{A}} \operatorname{M}_{\mathcal{A}}(T, \mathcal{F}) \leq R$. Since $\mathcal{B}$ is always guaranteed to observe a realizable sequence of labeled instances, picking the SOA \citep{Littlestone1987LearningQW, DanielyERMprinciple} gives that $R = \operatorname{L}(\mathcal{F})$. Since the SOA is deterministic, this choice of $\mathcal{B}$ implies that $\mathcal{A}$ is deterministic. Part (i) then follows from that the fact that for any deterministic learner $\mathcal{A}$ we have $\operatorname{M}_{\mathcal{A}}(T, \mathcal{F}) \geq \operatorname{C}_T(\mathcal{F})$. 
\end{proof}

\begin{proof} (of (iv) in Theorem \ref{thm:dimrel}) To prove part (iv), let $\mathcal{X} = [0, 1]$ and consider the class of thresholds $\mathcal{F} = \{x \mapsto \mathbbm{1}\{x \geq a\}: a \in (0, 1)\}$. It is well known that $\operatorname{L}(\mathcal{F}) = \infty$. On the other hand, note that $\dim(\mathcal{F}) = 1$ since $\mathcal{F}(1) = \{1\}$,  $\mathcal{F}(0) = \{0\}$, and $\mathcal{F}(x) = \{0, 1\}$ for all $x \in (0, 1)$.  
\end{proof}

\section{Proofs for Linear Systems}\label{app:linearsystems}

\begin{proof}(of (i) in Theorem \ref{thm:linsys})
We first prove the lower bound. It suffices to show that 

$$ \inf_{\text{Deterministic }\Acal }\,\, \operatorname{M}_{{\mathcal{A}}}(T, \mathcal{F}) \geq r+1.$$

Let $\Acal$ be any deterministic algorithm and $\{e_0, e_1, \ldots, e_r\}$ be arbitrary $r+1$ standard basis on $\reals^n$. This set exists because $n \geq r+1$. Consider a stream such that $x_0=e_0$,  $x_{t} = \{-e_{t}, e_{t}\} \backslash \Acal(x_{<t})$ for all $ t \in [r]$, and $  x_{r+1} = \{-e_1, e_1\} \backslash \Acal(x_{<r+1}).$
Here, $\forall t \in [r]$, we choose $x_t$ to be an element of the set $\{-e_{t}, e_{t}\}$ other than $\Acal$'s prediction on round $t$. Similarly, $x_{r+1}$ is chosen among $\{-e_1, e_1\}$.
We can define a stream using $\Acal$ because such a stream can be simulated before the game starts as $\Acal$ is deterministic. Let $\{\sigma_t\}_{t=0}^{r+1} \in \{-1,1\}^{r+2}$ such that $(x_0, \ldots, x_{r+1}) = (\sigma_0\, e_0, \sigma_1 e_1, \sigma_2 e_2, \ldots, \sigma_{r} e_r, \sigma_{r+1} e_1)$. Consider the integer-valued matrix $W^{\star} = \sum_{t=1}^{r} \sigma_{t-1}\sigma_{t} \, e_{t} \otimes e_{t-1}  + \sigma_r \sigma_{r+1} \,  e_1 \otimes e_r  $.
For $r+2\leq t \leq T$, one can define the stream to be $x_t = W^{\star}x_{t-1}$. Note that the $\text{image}(W^{\star}) \subseteq \text{span}(\{e_1, \ldots, e_r\})$. Thus, $\text{rank}(W^{\star}) \leq r$ and $W^{\star} \in \Fcal$. Finally, since $W^{\star}$ satisfies $W^{\star}(\sigma_{t-1} e_{t-1}) = \sigma_{t}\, e_{t}$ for $t \in [r]$,  $W^{\star}(\sigma_{r} e_r) = \sigma_{r+1} e_1$, and $W^{\star}x_{t-1} = x_t$ for $t\geq r+2$, the stream is realizable. By the definition of the stream, $\Acal$ makes mistakes on all rounds $t \in \{1, \ldots, r+1\}$. Thus, we have $\operatorname{M}_{{\mathcal{A}}}(T, \mathcal{F}) \geq r+1$.

To prove the upper bound, fix $T \in \mathbbm{N}$ and suppose that $T \geq r+1$ (otherwise $\operatorname{C}_T(\mathcal{F}) \leq T \leq r+1$). Consider a trajectory tree $\mathcal{T}$ of depth $T$ shattered by $\Fcal$. We first show that, for any path $\sigma \in \{-1, 1\}^T$ down $\mathcal{T}$ and $t \in [T-1]$, if 
 $\mathcal{T}_{t+1}((\sigma_{\leq t}, -1)) \neq \mathcal{T}_{t+1}((\sigma_{\leq t}, +1))$, then $\mathcal{T}_{t}(\sigma_{\leq t}) \notin \text{span}(\{\mathcal{T}_0, \mathcal{T}_1(\sigma_{\leq 1}),...,\mathcal{T}_{t-1}(\sigma_{\leq t-1}\})$. To prove the contraposition of this statement, suppose $\mathcal{T}_{t}(\sigma_{\leq t}) \in \text{span}(\{\mathcal{T}_0, \mathcal{T}_1(\sigma_{\leq 1}),...,\mathcal{T}_{t-1}(\sigma_{\leq t-1}\}) $. Then, there exists constants $a_0, ..., a_{t-1} \in \mathbbm{Z}$ such that $a_0 \mathcal{T}_0 + ..., + a_{t-1}\mathcal{T}_{t-1}(\sigma_{\leq t-1}) = \mathcal{T}_{t}(\sigma_{\leq t}) $. Thus, every matrix $W$ consistent with the sequence $\mathcal{T}_0, \mathcal{T}_1(\sigma_{\leq 1}),...,\mathcal{T}_{t-1}(\sigma_{\leq t-1})$ satisfies $W \Tcal_{t}(\sigma_{\leq t}) = W(a_0 \mathcal{T}_0 + ..., + a_{t-1}\mathcal{T}_{t-1}(\sigma_{\leq t-1}) ) = a_0 \mathcal{T}_1(\sigma_{\leq 1}) + ..., + a_{t-1}\mathcal{T}_{t}(\sigma_{\leq t})$. This implies that the path $\sigma_{\leq t+1}$ is shattered only if $\mathcal{T}_{t+1}((\sigma_{\leq t}, -1)) = \mathcal{T}_{t+1}((\sigma_{\leq t}, +1))$. 

Next, we show that the branching factor of every path in $\Tcal$ can be at most $r+1$. Suppose, for the sake of contradiction, there exists a path $\sigma \in \{-1,1\}^{T}$ whose branching factor is  $k > r+1$. We are guaranteed an increasing sequence of time points $t_1, t_2, \ldots, t_k $  in $\{0, \ldots, T-1\}$ such that  $\mathcal{T}_{t_i+1}((\sigma_{\leq t_i}, -1)) \neq \mathcal{T}_{t_i+1}((\sigma_{\leq t_i}, +1))$ for all $i \in [k]$. Since $t_2 \geq 1 $, by our argument above, we are guaranteed that $ \Tcal_{t_i}(\sigma_{\leq t_i}) \notin \text{span}(\{\Tcal_0, \ldots, \Tcal_{t_i-1}(\sigma_{\leq t_i-1})\})$ for all $i \in \{2, \ldots, k\}$. This further implies that the set $S := \{\Tcal_{t_2}(\sigma_{\leq t_2}), \ldots, \Tcal_{t_k}(\sigma_{\leq t_k})\}$ is a linearly independent set. Note that $|S| = k-1 > r$. Let $W_{\sigma}$ be the matrix such that the associated function $f_{\sigma} \in \Fcal$ shatters this path $\sigma$. By the definition of the tree, we must have that $S \subseteq \text{image}(W_{\sigma})$. Since $\text{rank}(W_{\sigma}) \leq r$, the set $\text{image}(W_{\sigma})$ must be a subset of a vectorspace of dimension  $\leq r$. However, this contradicts the fact that $S$ contains at least $r+1$ linearly independent vectors in $\reals^n$. Thus, the branching factor of every path in $\Tcal$ is at most $r+1$ and $\operatorname{B}(\Tcal) \leq r + 1$. Since $\Tcal$ is arbitrary, we have $\operatorname{C}_T(\Fcal) \leq r+1$. \end{proof}

\begin{proof}(of (ii) in Theorem \ref{thm:linsys}) We claim that it is without loss of generality to consider the function class $\mathcal{G} = \{x \mapsto Mx\, (\text{mod } 2)\, :\, M \in \{0, 1\}^{n \times n}\} \subset \mathcal{F}$. Indeed, for every $f \in \mathcal{F}$, there exists a $g \in \mathcal{G}$ such that $f(x) = g(x)$ for all $x \in \Xcal$. To see this, let $W \in \integers^{n \times n}$ be such that $f(x) = Wx (\text{mod } 2)$ and fix some $x \in \{0, 1\}^n$. Then, observe that $f(x) = Wx (\text{mod } 2) = (W(\text{mod } 2) x) (\text{mod } 2) = Mx(\text{mod } 2)$ where $M = W(\text{mod } 2) \in \{0, 1\}^{n \times n}$. Thus the function $g \in \mathcal{G}$ parameterized by $M$ matches $f$ everywhere on $\Xcal$. We now prove that $\operatorname{C}_T(\mathcal{G}) = n$. In the lower bound, we will use the fact that for a system of $n$ linearly independent equations with $n$ free variables defined over the integers  modulo $2$, there exists a unique solution.

Starting with the upper bound, fix $T \in \mathbbm{N}$ and suppose that $T \geq n$ (as otherwise $\operatorname{C}_T(\mathcal{G}) \leq T \leq n$). Consider a trajectory tree $\mathcal{T}$ of depth $T$ shattered by $\mathcal{G}$. Let $\sigma \in \{-1, 1\}^T$ denote an arbitrary path down $\Tcal$. We first claim that for $t \geq 1$ if $\mathcal{T}_t(\sigma_{\leq t})$ is linearly dependent modulo $2$ on the preceding sequence of states $\Tcal_0, \Tcal_1(\sigma_{\leq 1}), ..., \Tcal_{t-1}(\sigma_{< t})$, then $\mathcal{T}_{t+1}((\sigma_{\leq t}, -1)) = \mathcal{T}_{t+1}((\sigma_{\leq t}, +1)).$ To see this, suppose that $\mathcal{T}_t(\sigma_{\leq t})$ is linearly dependent modulo $2$ on the preceding sequence of states $\Tcal_0, \Tcal_1(\sigma_{\leq 1}), ..., \Tcal_{t-1}(\sigma_{< t})$. Then, there exists constants $a_0, ..., a_{t-1} \in \{0, 1\}$ such that $a_0\Tcal_0 + ... + a_{t-1}\Tcal_{t-1}(\sigma_{< t}) (\text{mod } 2) = \mathcal{T}_t(\sigma_{\leq t}) .$ Thus, every function $g \in \mathcal{G}$ consistent with the sequence $\Tcal_0, \Tcal_1(\sigma_{\leq 1}), ..., \Tcal_{t-1}(\sigma_{< t})$ outputs $a_0\Tcal_1(\sigma_{\leq 1}) + ... + a_{t-1}\Tcal_{t}(\sigma_{\leq t})(\text{mod } 2)$ on  $\mathcal{T}_t(\sigma_{\leq t})$, implying that the paths $(\sigma_{\leq t}, -1)$ and $(\sigma_{\leq t}, +1)$ are shattered only if $\mathcal{T}_{t+1}((\sigma_{\leq t}, -1)) = \mathcal{T}_{t+1}((\sigma_{\leq t}, +1)) = a_0\Tcal_1(\sigma_{\leq 1}) + ... + a_{t-1}\Tcal_{t}(\sigma_{\leq t})(\text{mod } 2).$ As a consequence, for $t \geq 1$, if $\mathcal{T}_{t+1}((\sigma_{\leq t}, -1)) \neq \mathcal{T}_{t+1}((\sigma_{\leq t}, +1))$, then $\mathcal{T}_t(\sigma_{\leq t})$ is linearly independent modulo $2$ of its preceding states.  Next, we claim that there can be at most $n-1$ timepoints $t_1, ..., t_{n-1} \in [T-1]$ such that $\Tcal_{t_i}(\sigma_{\leq t_i})$ is linearly independent of its preceeding sequence of states $\Tcal_0, \Tcal_1(\sigma_{\leq 1}), ..., \Tcal_{t_i-1}(\sigma_{< t_i})$. Indeed, suppose for sake of contradiction there exists $n$ timepoints $t_1, ..., t_{n} \in [T-1]$ such that $\Tcal_{t_i}(\sigma_{\leq t_i})$ is linearly independent of its preceeding states. Then, note that the following set of $n+1$ states  $\{\mathcal{T}_0, \mathcal{T}_{t_1}(\sigma_{\leq t_1}), ..., \mathcal{T}_{t_n}(\sigma_{\leq t_n})\}$ are linearly independent modulo $2$. This is a contradiction since $\mathcal{X}$ is $n$-dimensional. Combining the two claims, we get that 

\begin{align*}
\sum_{t=1}^{T} \mathbbm{1}\{\mathcal{T}_t((\sigma_{<t}, -1)) \neq  \mathcal{T}_t\big((\sigma_{<t}, +1)\big) \} &\leq 1  +  \sum_{t=2}^T \mathbbm{1}\{\mathcal{T}_t((\sigma_{<t}, -1)) \neq  \mathcal{T}_t\big((\sigma_{<t}, +1)\big) \} \\
&\leq 1  +  \sum_{t=1}^{T-1} \mathbbm{1}\{\mathcal{T}_t(\sigma_{\leq t}) \text{ linearly indep. of preceeding states} \}\\
&\leq 1 + n - 1 = n. 
\end{align*}

Since $\sigma \in \{-1, 1\}^T$ is arbitrary, we get that $\operatorname{B}(\Tcal) \leq n$. Finally, since $\Tcal$ is arbitrary, we have that $\operatorname{C}_T(\Fcal) \leq n$ which completes the proof of the upper bound.

To prove the lower bound, let $\mathcal{A}$ be any deterministic algorithm. Let $\mathcal{E} = \{e_0, ..., e_{n-1}\} \subset \Xcal$ be the standard basis over $\mathbb{R}^{n}$. Consider the stream where we pick $x_0 = e_0$, for all $t \in [n-2]$,  we pick $x_t \in \mathcal{E} \setminus \{x_0, ..., x_{t-1},\Acal(x_{<t}) \}$. Let $e = \mathcal{E} \setminus \{x_0, ..., x_{n-2}\}$ be the remaining basis. Pick $x_{n-1} \in \{e , e + e_0\} \setminus \{\mathcal{A}(x_{<n-1})\}$. Finally, pick $x_n \neq \mathcal{A}(x_{<n})$. We can define a stream using $\Acal$ because it is deterministic and thus can be simulated before the game begins. Next, we show that $x_0, ..., x_n$ is a realizable stream. This follows from the fact that $x_0, ..., x_{n-1}$ are linearly independent by definition, and thus there exists a function $g \in \mathcal{G}$ such that $g(x_{t-1}) =x_t $ for all $t \in [n]$. Moreover, by definition of the stream, $\mathcal{A}$ makes a mistake in every round. Thus, $\operatorname{M}_{\Acal}(T, \Fcal) \geq n$. Using the fact that $\operatorname{M}_{\Acal}(T, \Fcal) \leq \operatorname{C}_{T}(\mathcal{F})$ completes the proof. 
\end{proof}

\begin{proof}(of (iii) in Theorem \ref{thm:linsys})
Starting with the lower bound, it suffices to show that
$$\inf_{\text{Deterministic }\Acal } \operatorname{M}_{{\mathcal{A}}}(T, \mathcal{F}) \geq n.$$
Let $\Acal$ be any deterministic algorithm and $\{e_1, \ldots, e_n\}$ be the standard basis on $\reals^n$. Consider a stream such that $x_0= e_1 + e_2+ \ldots + e_n $, then $x_1 = x_0-e_{i_1}$ for some $i_1 \in [n]$ such that $x_1 \neq \Acal(x_0)$,  and for all $t \in \{2, \ldots, n-1\}$,
\[x_{t} =  x_0 - (e_{i_1} + \ldots + e_{i_{t}}) \text{ such that } e_{i_t} \notin \{e_{i_1}, \ldots, e_{i_{t-1}}\} \text{ and } x_{t} \neq \Acal(x_{<t}) .\]
Define $e_{i_n} = x_0 - (e_{i_1} + \ldots + e_{i_{n-1}}) $. Note that $x_{n-1} = e_{i_n}$. Finally, we choose $x_n \in \{e_{i_n}, \mathbf{0}\}$ such that $x_n \notin \Acal(x_{< n-1})$. Here, $\forall t \in [n-2]$, we choose $x_t$ by subtracting a basis $e_{i_t}$, which has not been subtracted before, from $x_{t-1}$ such that $x_t$ is other than what $\Acal$ would have predicted on round $t$. Finally, for $x_n$, we either choose it to be equal to $ x_{n-1} = e_{i_n}$ again or $\mathbf{0}$ while ensuring that $x_n \neq \Acal(x_{<{n-1}})$.  We can define a stream using $\Acal$ because such a stream can be simulated before the game starts as $\Acal$ is deterministic.

Next, we show that $(x_0, \ldots, x_n)$ is a realizable stream. Indeed, the boolean matrix 
\[W^{\star} =\sum_{k=2}^{n} e_{i_k} \otimes (e_{i_1} + \ldots+ e_{i_{k-1}})  \, + x_n \otimes e_{i_n}.  \]
satisfies $\indicator\{W^{\star}x_{t-1} > 0\}=x_{t}$ for all $t \in [n]$. For $t = 1$, we have 
\begin{equation*}
    \begin{split}
     W^{\star}x_0 &= \left(\sum_{k=2}^{n} e_{i_k} \inner{e_{i_1}+ \ldots + e_{i_{k-1}}}{x_0} \right) + x_n\\
     &= \sum_{k=2}^{n}(k-1) e_{i_k} + x_n  \\
     &= (k-1)(x_0-e_{i_1})+x_n
    \end{split}
\end{equation*}

Thus, $\indicator\{W^{\star} x_{0} >0\} = x_0 -e_{i_1} = x_1$. For $t \in \{2, \ldots, n-1\}$, we have 
\begin{equation*}
    \begin{split}
        W^{\star}x_{t-1} &= \left(\sum_{k=2}^{n} e_{i_k} \inner{e_{i_1}+ \ldots + e_{i_{k-1}}}{x_0- \left(e_{i_1} + \ldots + e_{i_{t-1}}\right)} \right) + x_n\\
        &= \sum_{k=2}^n e_{i_k}\inner{e_{i_1}+ \ldots + e_{i_{k-1}}}{e_{i_{t}} + \ldots + e_{i_n}} + x_n\\
        &= \sum_{k=t+1}^{n} (k-t)\,  e_{i_k} + x_n. 
    \end{split}
\end{equation*}
So, we obtain  $\indicator\{W^{\star} x_{t-1} >0\} = x_0 -(e_{i_1}+ \ldots + e_{i_t})>0 = x_t$. Finally, since $x_{n-1} = e_{i_n}$ and $W^{\star} e_{i_n} = x_n$, we have  $\indicator\{W^{\star}x_{n-1} >0 \}= x_n$. By the definition of the stream, $\Acal$ makes mistakes in every round on this stream. Thus, $\operatorname{M}_{{\mathcal{A}}}(T, \mathcal{F}) \geq n$.

As for the upper bound, since $|\{0,1\}^{n \times n}| = 2^{n^2}$, we have $\operatorname{L}(\Fcal) \leq \log(2^{n^2}) = n^2$. Thus, by Theorem \ref{thm:dimrel} (iii), we must have $\operatorname{C}_{T}(\Fcal) \leq n^2$. 
\end{proof}

\section{Proof of Theorem \ref{thm:01agn}} \label{app:thm01agn}

\begin{proof} (of upper bound in Theorem \ref{thm:01agn})
Let $(x_0, x_1, ..., x_T)$ be the stream to be observed by the learner. Given a multiclass classification learner $\mathcal{B}$ for $\Fcal$, consider the algorithm $\Acal$ as defined in Algorithm \ref{alg:det_SOA}. Here, $\Acal$ makes the same predictions as $\mathcal{B}$ while simulating the stream of labeled instance $(x_0, x_1), ..., (x_{T-1}, x_T)$ to $\mathcal{B}$. Using the bound (i) in Equation \eqref{reduction}, we obtain
\[\mathbb{E}\left[\sum_{t=1}^{T}\indicator\{\hat{x}_t \neq x_t\}\right] \leq \inf_{f \in \Fcal} \sum_{t=1}^T \indicator\{f(x_{t-1}) \neq x_t\} + R,\]
where expectation is taken with respect to the randomness of $\mathcal{B}$ and $R$ is the expected regret of the $\mathcal{B}$. This shows that $ \operatorname{MR}_{\mathcal{A}}(T, \mathcal{F}) \leq R$. Taking $\mathcal{B}$ to be $\mathbb{A}_{\text{AG}}$ defined in \cite[Section 3]{hanneke2023multiclass}, Theorem 4 in \citep{hanneke2023multiclass} implies that $R \leq \operatorname{L}(\mathcal{F}) + \sqrt{T \, \operatorname{L}(\mathcal{F})\log T}$, matching the claimed upper bound in Theorem \ref{thm:01agn}.
\end{proof}

\begin{proof}(of lower bound of $\frac{\operatorname{L}(\Fcal)}{18}$ in Theorem \ref{thm:01agn}) Define $d := \operatorname{L}(\Fcal)$ and let $\Tcal$ be the Littlestone tree of depth $d$ shattered by $\Fcal$. Let $(\Tcal_0, \ldots, \Tcal_{d-1})$ be the sequence of node-labeling functions and $(Y_1, \ldots, Y_d)$ be the sequence of edge labeling functions of the shattered tree $\Tcal$.  To construct a stream, take $r = \floor{\frac{d+2}{3}}$ and define a random sequence $\{n_i\}_{i=1}^r$ such that 
\[n_1 = 1,  \quad n_{i+1} \sim \text{Unif}(\{3i-1, 3i, 3i+1\}) \text{ for all } i \in [r-1]. \]
 Note that $n_r \leq 3(r-1)+1 = 3r-2 \leq d$. Next, pick a random path $(\sigma_1, \ldots, \sigma_d)$ down the tree  such that $\sigma_i \sim \text{Unif}\{-1, 1\}$. Let $x_0= \Tcal_0$ be the initial state and consider the stream
\[Y_{n_1}(\sigma_{\leq n_1}), \Tcal_{n_2-1}(\sigma_{<n_2}), Y_{n_2}(\sigma_{\leq n_2}), \ldots, \Tcal_{n_r-1}(\sigma_{<n_r}), Y_{n_r}(\sigma_{\leq n_r}). \]


For any randomized algorithm $\Acal$, let $\Acal_t$ denote its prediction on round $t$. Then, its expected cumulative loss on this stream is
 \begin{equation*}
     \begin{split}
        &\mathbb{E}\left[\sum_{t=1}^r \indicator\{\Acal_t \neq Y_{n_t}(\sigma_{\leq n_t})\} + \sum_{t=2}^{r} \indicator\{\Acal_t \neq \Tcal_{n_{t}-1}(\sigma_{<n_{t}}) \} \right],
     \end{split}
 \end{equation*}
 where the expectation is taken with respect to $\Acal$, $\sigma$, and $n_t$'s.
 Note that 
 \begin{equation*}
     \begin{split}
       \expect[\indicator\{\Acal_t \neq Y_{n_t}(\sigma_{\leq n_t})\} ]
       \geq \frac{1}{2}.  
     \end{split}
 \end{equation*}
 where the inequality holds because conditioned on the history up to time point $t-1$, the state $ Y_{n_t}(\sigma_{\leq n_t})$ is chosen uniformly at random between $Y_{n_t}((\sigma_{<n_t}, -1))$ and $Y_{n_t}((\sigma_{<n_t}, +1))$. So, the algorithm cannot do better than random guessing. Similarly, given a path $\sigma$, the state $\Tcal_{n_{t}-1}(\sigma_{<n_{t}})$ is selected uniformly from the set
 \[\left\{\Tcal_{m-1}(\sigma_{<m})\, : m=3(t-1)-1, 3(t-1), 3(t-1)+1 \right\}.\]
Since each node along a path must have distinct elements, the aforementioned set must contain $3$ elements. Thus,  we have
 \[\expect[\indicator\{\Acal_t \neq \Tcal_{n_{t}-1}(\sigma_{<n_{t}}) \} ] \geq \frac{2}{3} .\]
 Overall, the total expected cumulative loss of $\Acal$ is 
 \[\geq  \sum_{t=1}^r \frac{1}{2} + \sum_{t=2}^{r} \frac{2}{3} \geq \frac{r}{2} + \frac{2(r-1)}{3} = \frac{7r-4}{6}.\]

To upper bound the loss of the best-fixed competitor, let $f_{\sigma}\in \Fcal$ denote the function that shatters the path $\sigma$ in the tree $\Tcal$. Then, by definition of shattering, we have  $f_{\sigma}(\Tcal_{n_{t}-1}(\sigma_{<n_{t}})) = Y_{n_t}(\sigma_{\leq n_t})$. Thus, the cumulative loss of $f_{\sigma}$ is 
\begin{equation*}
    \begin{split}
        \mathbb{E}\Bigg[\sum_{t=1}^r \indicator\{f_{\sigma}(\Tcal_{n_t-1}(\sigma_{<n_t})) \neq Y_{n_t}(\sigma_{\leq n_t})\} &+ \sum_{t=2}^{r} \indicator\{f_{\sigma}(Y_{n_{t-1}}(\sigma_{\leq n_{t-1}})) \neq \Tcal_{n_{t}-1}(\sigma_{<n_{t}}) \} \Bigg] \\
        &\leq 0 + \sum_{t=2}^{r} 1 = r-1.
    \end{split}
\end{equation*}
Hence, combining everything and plugging the value of $r$, the regret of $\Acal$ is 
\[\operatorname{MR}_{\Acal}(T, \Fcal) \geq \frac{7r-4}{6}-(r-1) = \frac{7r-4-6r+6}{6} = \frac{r+2}{6}\geq \frac{(d+2)/3-1+2}{6} \geq \frac{d}{18}.\]
This completes our proof.\end{proof}

\begin{proof}(of lower bound of $\frac{\sqrt{T}}{16\sqrt{3}}$ in Theorem \ref{thm:01agn}) Let $f_{-1}$ and $f_{+1}$ be any two distinct functions in $\Fcal$ and $\bar{x}_0 \in \Xcal$ be the point where they differ. That is, $f_{-1}(\bar{x}_0) \neq f_{+1}(\bar{x}_0)$. Define $S \subset \Xcal$ such that $f_{-1}(x) = f_{+1}(x)$ for all $x \in S$. Moreover, define $S_0 = \{x \in S \mid f_{-1}(x) = f_{+1}(x) = \bar{x}_0\}$. Let $(\sigma_1, \ldots, \sigma_T)$ be a sequence such that $\sigma_t \sim \text{Uniform}\{-1, 1\}$. Consider a random stream with initial state $x_0 = \bar{x}_0$ and for all $t \in [T]$,
\[x_{t} =\bar{x}_0 \indicator\{x_{t-1} \in S_0\} + \text{Uniform}(\{\bar{x}_0, f_{\sigma_t}(x_{t-1})\})\,  \indicator\{x_{t-1} \in S\backslash S_0\} + f_{\sigma_t}(x_{t-1}) \, \indicator\{x_{t-1} \notin S\}. \]

For any algorithm $\Acal$, its expected cumulative loss is 
\begin{equation*}
    \begin{split}
        \mathbb{E}\left[\sum_{t=1}^T \indicator\{\Acal(x_{<t}) \neq x_{t}\} \right] &= \mathbb{E}\left[\sum_{t=1}^T \indicator\{\Acal(x_{<t}) \neq x_{t}\} 
 \indicator\{x_{t-1} \in S_0\}\right] +  \mathbb{E}\left[\sum_{t=1}^T \indicator\{\Acal(x_{<t}) \neq x_{t}\} 
 \indicator\{x_{t-1} \notin S_0 \}\right]\\
 &\geq 0 \, + \,   \mathbb{E}\left[\sum_{t=1}^{T} \expect \big[  \indicator\{\Acal(x_{<t}) \neq x_{t}\} 
 \mid x_{<t}, \Acal \big] \, \indicator\{x_{t-1} \notin S_0 \} \right] \\
 &\geq  \frac{1}{2}\,\mathbb{E}\left[\sum_{t=1}^{T} \indicator\{x_{t-1} \notin S_0 \} \right],
    \end{split}
\end{equation*}
where the final step follows because when $x_{t-1} \notin S_0$, the state $x_t$ is sampled uniformly at random between two states. So, the expected loss of algorithm in this round is $\geq \frac{1}{2}$.

To upper bound the cumulative loss of the best-fixed function in hindsight, define $\sigma = \text{sign}\left( \sum_{t=1}^T \sigma_t \indicator\{x_{t-1} \notin S\}\right)$.
Note that
\begin{equation*}
    \begin{split}
       &\mathbb{E}\left[  \inf_{f \in \Fcal} \sum_{t=1}^T\indicator\{f(x_{t-1}) \neq x_t\}\right] \\
        &\leq \mathbb{E}\left[ \sum_{t=1}^T\indicator\{f_{\sigma}(x_{t-1}) \neq x_t\}\right] \\
        &= 0+ \mathbb{E}\left[ \sum_{t=1}^T   \indicator\{f_{\sigma}(x_{t-1}) \neq x_t\} \indicator\{x_{t-1} \in S\backslash S_0\}  \right] + \mathbb{E}\left[ \sum_{t=1}^T   \indicator\{f_{\sigma}(x_{t-1}) \neq f_{\sigma_t}(x_{t-1})\} \indicator\{x_{t-1} \notin S\}  \right]\\
        &\leq \frac{1}{2}\mathbb{E}\left[ \sum_{t=1}^T  \indicator\{x_{t-1} \in S\backslash S_0\}  \right]  + \mathbb{E}\left[ \sum_{t=1}^T   \indicator\{f_{\sigma}(x_{t-1}) \neq f_{\sigma_t}(x_{t-1})\} \indicator\{x_{t-1} \notin S\}  \right],
    \end{split}
\end{equation*}
where the final step follows because conditioned on the event that $x_{t-1} \in S\backslash S_0$, the state $x_{t}$ is $ \sim \text{Uniform}(\{\bar{x}_0, f_{\sigma_t}(x_{t-1})\})$. Since $f_{\sigma}(x_{t-1}) = f_{\sigma_t}(x_{t-1})$ whenever $x_{t-1} \in S$, the expected loss of $f_{\sigma}$ on round $t$ is equal to the conditional probability that $x_t = x_0$, which is $\leq 1/2$. Moreover, using the fact that
 $ \indicator\{f_{\sigma}(x_{t-1}) \neq f_{\sigma_t}(x_{t-1})\}  \leq \indicator\{\sigma \neq \sigma_t\}$, we have
 
\begin{equation*}
    \begin{split}
           \mathbb{E}\left[ \inf_{f \in \Fcal} \sum_{t=1}^T\indicator\{f(x_{t-1}) \neq x_t\}\right]&\leq \frac{1}{2}\mathbb{E}\left[ \sum_{t=1}^T  \indicator\{x_{t-1} \in S\backslash S_0\}  \right] + \mathbb{E}\left[ \sum_{t=1}^T   \indicator\{\sigma \neq \sigma_t\} \indicator\{x_{t-1} \notin S\}  \right]\\
           &= \frac{1}{2}\mathbb{E}\left[ \sum_{t=1}^T  \indicator\{x_{t-1} \in S\backslash S_0\}  \right] + \mathbb{E}\left[ \sum_{t=1}^T   \frac{1-\sigma\sigma_t}{2} \indicator\{x_{t-1} \notin S\}  \right]\\
           &= \frac{1}{2} \mathbb{E}\left[ \sum_{t=1}^T\indicator\{x_{t-1} \notin S_0\}\right] -\frac{1}{2}\mathbb{E}\left[ \sigma \sum_{t=1}^{T}\sigma_t \indicator\{x_{t-1} \notin S\}\right]\\
           &= \frac{1}{2} \mathbb{E}\left[ \sum_{t=1}^T\indicator\{x_{t-1} \notin S_0\}\right] -\frac{1}{2}\mathbb{E}\left[ \left| \sum_{t=1}^{T}\sigma_t \indicator\{x_{t-1} \notin S\} \right|\right],
    \end{split}
\end{equation*}
where the final equality follows from the definition of $\sigma$.
Thus, the regret of $\Acal$ is 
\begin{equation*}
    \begin{split}
        \operatorname{MR}_{\Acal}(T, \Fcal) &= \mathbb{E}\left[\sum_{t=1}^T \indicator\{\Acal(x_{<t}) \neq x_{t}\} \right] -     \mathbb{E}\left[ \inf_{f \in \Fcal} \sum_{t=1}^T\indicator\{f(x_{t-1}) \neq x_t\}\right] \geq \frac{1}{2} \mathbb{E}\left[ \left| \sum_{t=1}^{T}\sigma_t \indicator\{x_{t-1} \notin S\} \right|\right]. 
    \end{split}
\end{equation*}
We now lower bound the Rademacher sum above by closely following the proof of Khinchine's inequality \cite[Lemma 8.1]{cesa2006prediction}.

For any random variable $Y$ with the finite-fourth moment, a simple application of Holder's inequality implies that 
\[\mathbb{E}[|Y|] \geq \frac{(\mathbb{E}[Y^2])^{3/2}}{(\mathbb{E}[Y^4])^{1/2}}.\]
We apply this inequality to $Y := \sum_{t=1}^{T}\sigma_t \indicator\{x_{t-1} \notin S\}$. Since $Y^2 = \sum_{t=1}^{T}\indicator\{x_{t-1} \notin S\} + 2 \sum_{i < j} \sigma_i \sigma_j \indicator\{x_{i-1}, x_{j-1} \notin S\}$, we have 
\[\expect[Y^2] = \expect\left[\sum_{t=1}^T \indicator\{x_{t-1} \notin S\}\right] +2\expect\left[ \sum_{i < j}  \expect[\sigma_j \mid \sigma_{<j}, x_{<j}] \, \, \sigma_i\indicator\{x_{i-1}, x_{j-1} \notin S\} \right]  = \expect\left[\sum_{t=1}^T \indicator\{x_{t-1} \notin S\}\right], \]
where the final equality follows because $\sigma_j$ is still a Rademacher random variable conditioned on the past and $\expect[\sigma_j \mid \sigma_{<j}, x_{<j}] =0$. Furthermore, note that 
\begin{equation*}
    \begin{split}
        \expect\left[\sum_{t=1}^T \indicator\{x_{t-1} \notin S\}\right] &\geq \expect\left[ \indicator\{x_{0} \notin S\} \right] + \sum_{r=1}^{\floor{\frac{T}{2}}-1} \expect\left[ \indicator\{x_{2r-1} \notin S\} + \indicator\{x_{2r} \notin S\}\right] \\
        &\geq 1 + \sum_{r=1}^{\floor{\frac{T}{2}}-1} \frac{1}{2}\\
        &= \frac{1}{2} \left( \floor*{\frac{T}{2}} +1 \right)\\
        &\geq \frac{T}{4}.
    \end{split}
\end{equation*}
Here, the second inequality above uses (i) $x_0 = \bar{x}_0 \notin S$ and (ii)  $\expect\left[ \indicator\{x_{2r-1} \notin S\} + \indicator\{x_{2r} \notin S\}\right]\geq \frac{1}{2}$ for all $r \geq 1$. To see (ii), one can consider two cases. First, if $x_{2r-1} \notin S$, then the inequality is trivially true. On the other hand, $x_{2r-1} \in S$, then $x_{2r}= \bar{x}_0 \notin S $ with probability $1/2$. Thus, we have $\mathbb{E}[Y^2] \geq \frac{T}{4}$.

Similarly, for the fourth moment of $Y$,  all the cross-term vanishes and we are left with the terms with fourth power and symmetric second powers. 
\begin{equation*}
    \begin{split}
        \expect[Y^4] &= \expect\left[\sum_{t=1}^T \indicator\{x_{t-1} \notin S\}\right] + \binom{4}{2}\expect\left[\sum_{i<j} \indicator\{x_{i-1}\notin S\}\, \indicator\{x_{j-1}\notin S\}\right]\\
        &\leq T + \binom{4}{2} \frac{T(T-1)}{2}\\
        &\leq T + 3T(T-1)\\
        &\leq 3T^2.
    \end{split}
\end{equation*}
The lower bound on the second moment of $Y$ and the upper bound on the fourth moment of $Y$ collectively implies that  
\[\mathbb{E}[|Y|] \geq  \frac{(\mathbb{E}[Y^2])^{3/2}}{(\mathbb{E}[Y^4])^{1/2}} \geq \frac{(T/4)^{3/2}}{(3T^2)^{1/2}} = \frac{\sqrt{T}}{8 \sqrt{3}}.\]
Finally, combining everything, we obtain
\[\operatorname{MR}_{\Acal}(T, \Fcal)  \geq \frac{1}{2} \mathbb{E}\left[ \left| \sum_{t=1}^{T}\sigma_t \indicator\{x_{t-1} \notin S\} \right|\right]=\frac{1}{2} \mathbb{E}[|Y|] = \frac{\sqrt{T}}{16\sqrt{3}}. \]
This completes our proof.
\end{proof}

\section{Tightness of upper bound in Theorem \ref{thm:01agn}} \label{app:tight_up_thm01agn}
\noindent Fix $d \in \mathbb{N}$ and let $\Xcal = \{1,2,\ldots, d\} \cup \{\pm (d+1), \pm (d+2), \ldots, \pm 2d\}$. For every $\sigma \in \{-1,1\}^d$, define a function $f_{\sigma}: \Xcal \to \Xcal$ such that
\[f_{\sigma}(x) = \sigma_x(d+x) \indicator\{x \in [d] \} + (|x|-d) \indicator\{ x \notin [d]\}.\]
Consider $\Fcal = \{f_{\sigma} \, :\, \sigma \in \{-1, 1\}^d\}$. It is not too hard to see that $\operatorname{L}(\Fcal) = d$. The fact that $\operatorname{L}(\Fcal) \leq d$ is trivial because any Littlesone tree can only have $\{1,2, \ldots, d\}$ in the internal nodes. This is because all functions output the same state in the domain $\Xcal \backslash [d]$. Since the internal nodes in any given path of the Littlestone tree need to be different, the depth of any shattered tree is $\leq d$. To see why $\operatorname{L}(\Fcal) \geq d$, consider a complete binary tree $\Tcal$ of depth $d$ with all the internal nodes in level $i \in [d]$ containing the element $i$. Let $-(d+i)$ and $(d+i)$ label the left and right outgoing edges respectively of every node in level $i$. Note that $\Tcal$ is shattered by $\Fcal$ as for any path $\epsilon \in \{-1, 1\}^d$ down the tree $\Tcal$, there exists a $f_{\epsilon} \in \Fcal$ such that $f_{\epsilon}(i) = \epsilon_i(d+i)$ for all $i \in [d]$. Thus, we have shown that $\operatorname{L}(\Fcal) = d$.

We now show that $\inf_{\mathcal{A}}\operatorname{MR}_{\mathcal{A}}(T, \mathcal{F}) = \Omega(\sqrt{T \operatorname{L}(\Fcal)})$, proving that the upper bound in Theorem \ref{thm:01agn} is tight up to $\sqrt{\log{T}}$. For a $k \in \naturals$, pick $T = 2kd-1$. Draw $ \varepsilon \in \{-1,+1\}^{T}$ where $\varepsilon_i \sim \text{Uniform}(\{-1, 1\})$ and consider a stream 
\begin{equation*}
    \begin{split}
    x_{2k(i-1)}= i, \,  x_{t} = \varepsilon_{t} (d +i) \indicator\{x_{t-1} =i\} + i\indicator\{x_{t-1} \neq i \} \quad \forall i \in [d] \text{ and } t \in \{2k(i-1)+1,  \ldots, 2ki-1\}. 
    \end{split}
\end{equation*}
Note that $x_0 = 1$ is the initial state and there are $T= 2kd-1$ more states in this stream. 
For any algorithm $\Acal$, its expected cumulative loss is
\begin{equation*}
    \begin{split}
        \mathbb{E}\left[ \sum_{t=1}^T \indicator\{\Acal(x_{<t}) \neq x_t\} \right] &\geq \mathbb{E}\left[ \sum_{i=1}^{d} \sum_{t=2k(i-1)+1}^{2ki-1} \indicator\{\Acal(x_{<t}) \neq x_t\} \right] 
        \geq   \frac{1}{2}
        \sum_{i=1}^{d}  \sum_{t=2k(i-1)+1}^{2ki-1} \indicator\{x_{t-1} = i\}.
    \end{split}
\end{equation*}
Here, the first inequality is true because we just rewrite the sum without including the rounds $t = 2k,4k, 6k,\ldots, 2(d-1)k$. The second inequality holds because $x_{t}$ is sampled uniformly at random between $-(d+i)$ and $d+i$ whenever $x_{t-1} = i$, and the algorithm cannot do better than guessing on these rounds. Note that there is no expectation following the second inequality because the event $x_{t-1}=i$ is deterministic. 

To upper bound the loss of the best function in hindsight, define $\bar{\varepsilon}_i = \text{sign}\big(\sum_{t=2k(i-1)+1}^{2ki-1} \varepsilon_t \indicator\{x_{t-1} = i\} \big)$ and consider $f_{\bar{\varepsilon}}$ where $\bar{\varepsilon} = (\bar{\varepsilon}_1, \ldots, \bar{\varepsilon}_d  )$. Then,
\begin{equation*}
    \begin{split}
           \mathbb{E}\left[  \inf_{f \in \Fcal} \sum_{t=1}^T\indicator\{f(x_{t-1}) \neq x_t\}\right] &\leq \mathbb{E}\left[ \sum_{t=1}^T\indicator\{f_{\bar{\varepsilon}}(x_{t-1}) \neq x_t\}\right] \\
        &\leq (d-1) + \mathbb{E}\left[ \sum_{i=1}^{d} \sum_{t=2k(i-1) +1 }^{2ki-1} \indicator\{f_{\bar{\varepsilon}}(x_{t-1}) \neq x_t\}\right] \\
        &= (d-1) + \mathbb{E}\left[ \sum_{i=1}^{d} \sum_{t=2k(i-1) +1 }^{2ki-1} \indicator\{f_{\bar{\varepsilon}}(x_{t-1}) \neq x_t\} \indicator\{x_{t-1} =i\} \right]. \\    
    \end{split}
\end{equation*}
Here, the second inequality holds because we trivially upper bound the losses on rounds $t=2k, 4k, \ldots, 2(d-1) k$ by $1$. And the final equality follows because 
$f_{\bar{\varepsilon}}(x_{t-1}) = i= x_t  \text{ whenever } x_{t-1} \neq i \text{ for } 2k(i-1)+1\leq t \leq 2ki-1$. Moreover, when $x_{t-1}=i$, we have $f_{\bar{\varepsilon}}(x_{t-1}) = \bar{\varepsilon}_i(d+i) $ and $x_t = \varepsilon_t (d+i)$. This implies that  $\indicator\{f_{\bar{\varepsilon}}(x_{t-1}) \neq x_t\} \indicator\{x_{t-1} =i\} =\indicator\{\bar{\varepsilon}_i \neq \varepsilon_t\} \indicator\{x_{t-1} =i\} $. Thus, we can write
\begin{equation*}
    \begin{split}
        \mathbb{E}\Bigg[  \inf_{f \in \Fcal} &\sum_{t=1}^T\indicator\{f(x_{t-1}) \neq x_t\}\Bigg]
        \\
        &\leq  (d-1) + \mathbb{E}\left[ \sum_{i=1}^{d} \sum_{t=2k(i-1) +1 }^{2ki-1} \indicator\{\bar{\varepsilon}_i \neq \varepsilon_t\} \indicator\{x_{t-1} =i\} \right]\\
        &= (d-1) + \mathbb{E}\left[ \sum_{i=1}^{d} \sum_{t=2k(i-1) +1 }^{2ki-1} \frac{1-\bar{\varepsilon}_i  \varepsilon_t}{2}\,  \indicator\{x_{t-1} =i\} \right]\\
        &= (d-1) +  \sum_{i=1}^{d} \sum_{t=2k(i-1) +1 }^{2ki-1} \frac{1}{2}\,  \indicator\{x_{t-1} =i\}  - \frac{1}{2}\mathbb{E}\left[ \sum_{i=1}^{d} \bar{\varepsilon}_i \sum_{t=2k(i-1) +1 }^{2ki-1}  \varepsilon_t \indicator\{x_{t-1} =i\} \right] \\
        &= (d-1) +  \sum_{i=1}^{d} \sum_{t=2k(i-1) +1 }^{2ki-1} \frac{1}{2}\,  \indicator\{x_{t-1} =i\}  - \frac{1}{2}\mathbb{E}\left[ \sum_{i=1}^{d} \left|\sum_{t=2k(i-1) +1 }^{2ki-1}  \varepsilon_t \indicator\{x_{t-1} =i\} \right|\right] \\
    \end{split}
\end{equation*}

Thus, the expected regret of $\Acal$ is 
\begin{equation*}
    \begin{split}
        \operatorname{MR}_{\mathcal{A}}(T, \mathcal{F}) &= \mathbb{E}\left[ \sum_{t=1}^T \indicator\{\Acal(x_{<t}) \neq x_t\} \right] -  \mathbb{E}\left[  \inf_{f \in \Fcal} \sum_{t=1}^T\indicator\{f(x_{t-1}) \neq x_t\}\right] \\
       &\geq \frac{1}{2}\mathbb{E}\left[ \sum_{i=1}^{d} \left|\sum_{t=2k(i-1) +1 }^{2ki-1}  \varepsilon_t \indicator\{x_{t-1} =i\} \right|\right] - (d-1)\\
       &\geq \frac{1}{2} \sum_{i=1}^d \sqrt{\frac{|\{ 2k(i-1)+1\leq t \leq 2ki-1 \, :\, x_{t-1} = i\}|}{2}} - (d-1)\\
       &= \frac{1}{2}\sum_{i=1}^d \sqrt{\frac{k}{2}} - (d-1).
    \end{split}
\end{equation*}
where the second inequality follows due to Khinchine's inequality \cite[Lemma 8.2]{cesa2006prediction}. The final equality holds because in each block of $2k$ rounds from $t=2k(i-1)+1$ to $t=2ki-1$, we have $x_{t-1}=i$ in exactly $k$ of those rounds. Using $T = 2kd-1$, we further obtain $ \operatorname{MR}_{\mathcal{A}}(T, \mathcal{F}) \geq \frac{1}{4}\sqrt{(T+1)d} - (d-1) = \Omega(\sqrt{Td})$ for $T \gg d$. This establishes our claim that the upper bound in Theorem \ref{thm:01agn} is tight up to $\sqrt{\log{T}}$.

\section{Proof of Theorem \ref{thm:02agn}} \label{app:prfthmagn}

 The lower bound in Theorem \ref{thm:02agn} follows directly from the lower bound in the realizable setting. Accordingly, we only prove the upper bound. Let $K := \sup_{x \in \Xcal} |\Fcal(x)|.$

\begin{proof} (of the upper bound in Theorem \ref{thm:02agn}) Let $\{x_t\}_{t=0}^T$ be the trajectory to be observed by the learner and $f^{\star} \in \argmin_{f \in \mathcal{F}} \sum_{t=1}^T \mathbbm{1}\{x_t \neq f^{t}(x_0)\}$ the optimal evolution function in hindsight. Given the time horizon $T$, let $L_T = \{L \subset [T]: |L| \leq \operatorname{C}_T(\mathcal{F})\}$ denote the set of all possible subsets of $[T]$ with size at most $\operatorname{C}_T(\mathcal{F})$. For every $L \in L_T$, let $\phi: L \rightarrow [K]$ denote a function mapping time points in $L$ to an integer in $[K]$. On time point $t \in L$, we should think of $\phi(t)$ as an index into the set $\mathcal{F}(x_{t-1})$. To that end, for an index $i \in [|\mathcal{F}(x)|]$, let $\mathcal{F}_i(x)$ denote the $i$-{th} element of the list obtained after sorting $\mathcal{F}(x)$ in its natural order.  Let $\Phi_L = [K]^L$ denote all such functions $\phi$. For each $L \in L_T$ and $\phi \in \Phi_L$, define an expert $\mathcal{E}_{L, \phi}$. On time point $t \in [T]$, the prediction of expert $\mathcal{E}_{L, \phi}$ is defined recursively by

$$\mathcal{E}_{L, \phi}(x_0, t) = \begin{cases}
			\mathcal{A}\left(x_0, t| \{\mathcal{E}_{L, \phi}(x_0, i)\}_{i=1}^{t-1} \right), & \text{if $t \notin L$}\\
            \mathcal{F}_{\phi(t)}(\mathcal{E}_{L, \phi}(x_0, t-1)), & \text{otherwise}
		 \end{cases}$$

where $\mathcal{E}_{L, \phi}(x_0, 0) = x_0$ and $\mathcal{A}\left(x_0, t| \{\mathcal{E}_{L, \phi}(x_0, i)\}_{i=1}^{t-1} \right)$ denotes the prediction of Algorithm \ref{alg:real} on timepoint $t$ after running and updating on the trajectory $\{\mathcal{E}_{L, \phi}(x_0, i)\}_{i=1}^{t-1}$. Let $E = \bigcup_{L \in L_T} \bigcup_{\phi \in \Phi_L} \mathcal{E}_{L, \phi}$ denote the set of all experts parameterized by $L \in L_T$ and $\phi \in \Phi_L$. 

We claim that there exists an expert $\mathcal{E}_{L^{\star}, \phi^{\star}}$ such that $\mathcal{E}_{L^{\star}, \phi^{\star}}(x_0, t) = f^{\star, t}(x_0)$ for all $t \in [T]$. To see this, consider the hypothetical trajectory $S^{\star} = \{f^{\star, t}(x_0)\}_{t=1}^T$  generated by $f^{\star}$. Let $L^{\star} = \{t_1, t_2, ...\}$ be the indices on which Algorithm \ref{alg:real} would have made a mistake had it run and updated on $S^{\star}$. By the guarantee of Algorithm \ref{alg:real}, we know that $|L^{\star}| \leq \operatorname{C}_T(\mathcal{F})$. Moreover, there exists a $\phi^{\star} \in \Phi_{L^{\star}}$ such that $\mathcal{F}_{\phi^{\star}(i)}(f^{\star, i-1}(x_0)) = f^{\star, i}(x_0)$ for all $i \in L^{\star}$. By construction of $\mathcal{E}$, there exists an expert $\mathcal{E}_{L^{\star}, \phi^{\star}}$ parameterized by $L^{\star}$ and $\phi^{\star}$. We claim that $\mathcal{E}_{L^{\star}, \phi^{\star}}(x_0, t) = f^{\star, t}(x_0)$ for all $t \in [T]$. 
This follows by strong induction on $t \in [T]$. For the base case $t = 1$, there are two subcases to consider. If $1 \in  L^{\star}$, then we have that $\mathcal{E}_{L^{\star}, \phi^{\star}}(x_0, 1) = \mathcal{F}_{\phi^{\star}(1)}(\mathcal{E}_{L^{\star}, \phi^{\star}}(x_0, 0)) = \mathcal{F}_{\phi^{\star}(1)}(x_0) = f^{\star, 1}(x_0)$. If $1 \notin L^{\star}$, then $\mathcal{E}_{L^{\star}, \phi^{\star}}(x_0, 1) = \mathcal{A}\left(x_0, 1| \{\} \right) = f^{\star, 1}(x_0)$, where the last equality follows by definition of $L^{\star}$. Now for the induction step, suppose that $\mathcal{E}_{L^{\star}, \phi^{\star}}(x_0, i) = f^{\star, i}(x_0)$ for all $i \leq t$. Then, if $t+1 \in L^{\star}$, we have that $\mathcal{E}_{L^{\star}, \phi^{\star}}(x_0, t+1) = \mathcal{F}_{\phi^{\star}(t+1)}(\mathcal{E}_{L^{\star}, \phi^{\star}}(x_0, t)) = \mathcal{F}_{\phi^{\star}(t+1)}(f^{\star, t}(x_0)) = f^{\star, t+1}(x_0).$ If $t+1 \notin L^{\star}$, then $\mathcal{E}_{L^{\star}, \phi^{\star}}(x_0, t+1) = \mathcal{A}\left(x_0, t+1| \{\mathcal{E}_{L^{\star}, \phi^{\star}}(x_0, i)\}_{i=1}^{t} \right) = \mathcal{A}\left(x_0, t+1| \{f^{\star, i}(x_0)\}_{i=1}^{t} \right) = f^{\star, t+1}(x_0)$, where the last equality again is due to the definition of $L^{\star}$. 

Now, consider the learner that runs the celebrated Randomized Exponential Weights Algorithm, denoted hereinafter by $\mathcal{P}$, using the set of experts $E$ with learning rate $\eta = \sqrt{\frac{2 \ln(|E|)}{T}}$. By Theorem 21.11 of \cite{ShwartzDavid}, we have that

\begin{align*}
\mathbbm{E}\left[\sum_{t=1}^T \mathbbm{1}\{\mathcal{P}(x_0, t) \neq x_t\} \right] &\leq \inf_{\mathcal{E} \in E} \sum_{t=1}^T \mathbbm{1}\{\mathcal{E}(x_0, t) \neq x_t\} + \sqrt{2T \ln(|E|)}\\
&\leq \inf_{\mathcal{E} \in E} \sum_{t=1}^T \mathbbm{1}\{\mathcal{E}(x_0, t) \neq x_t\} + \sqrt{2 \operatorname{C}_T(\mathcal{F}) \, T \ln\Bigl(\frac{KT}{\operatorname{C}_T(\mathcal{F})}\Bigl)}\\
&\leq  \sum_{t=1}^T \mathbbm{1}\{\mathcal{E}_{L^{\star}, \phi^{\star}}(x_0, t) \neq x_t\} + \sqrt{2 \operatorname{C}_T(\mathcal{F}) \, T \ln\Bigl(\frac{KT}{\operatorname{C}_T(\mathcal{F})}\Bigl)}\\
&= \sum_{t=1}^T \mathbbm{1}\{f^{\star, t}(x_0) \neq x_t\} + \sqrt{2 \operatorname{C}_T(\mathcal{F}) \, T \ln\Bigl(\frac{KT}{\operatorname{C}_T(\mathcal{F})}\Bigl)}
\end{align*}

where the second inequality follows because $|E| = \sum_{i=0}^{\operatorname{C}_T(\mathcal{F})} K^i \, \binom{T}{i} \leq \Bigl(\frac{KT}{\operatorname{C}_T(\mathcal{F})} \Bigl)^{\operatorname{C}_T(\mathcal{F})}$. This completes the proof as $\mathcal{P}$ achieves the stated upper bound on expected regret. \end{proof}

The next two examples show that both the lower- and upper bounds in Theorem \ref{thm:02agn} can be tight. 

\subsection{Tightness of lower bound in Theorem \ref{thm:02agn}} \label{exm: agnlbtight}
 \noindent  Let $\mathcal{X} = \mathbb{Z}$ and fix $p \in \mathbb{N}$.  For every $\sigma \in \{-1, 1\}^{\mathbb{N} \cup \{0\}}$, define the evolution function 

$$
f_{\sigma}(x) = (\sigma_{|x|}\mathbbm{1}\{|x| \leq p-1\} + \mathbbm{1}\{|x| \geq p\}) (|x|+1)
$$

\noindent and consider the class $\mathcal{F} = \Bigl\{f_{\sigma}: \sigma \in \{-1, 1\}^{\mathbb{N} \cup\,  \{0\}}\Bigl\}.$ By construction of $\mathcal{F}$, branching only occurs on states in $\{0, 1, ..., p-1\}$ and their negation. Moreover, given any initial state $x_0 \in \mathcal{X}$ and a time horizon $T \in \mathbb{N}$, the trajectory of any evolution in $\mathcal{F}$ is some signed version of the sequence $|x_0| + 1, ..., |x_0| + T$. From Theorem \ref{thm:everyrate}, its not too hard to see that $\operatorname{C}_T(\mathcal{F}) = \min\{p, T\}$.  We now show that $\inf_{\mathcal{A}}\operatorname{FR}_{\mathcal{A}}(T, \mathcal{F}) \leq \frac{\operatorname{C}_T(\mathcal{F})}{2}$. Consider the learner $\mathcal{A}$ which, given initial state $x_0$, checks whether $|x_0| \leq p-1$. If $|x_0| \leq p-1$, the learner $\mathcal{A}$ samples a random sequence $\epsilon \sim \{-1, 1\}^{p - |x_0|}$ and plays $\epsilon_t (|x_{0}| + t)$ for $t \leq p - |x_0|$ and $|x_{0}| + t$ in all future rounds $t > p - |x_0|$. If $|x_0| > p - 1$, the learner $\mathcal{A}$ plays $|x_{0}| + t$ for all $t \geq 1$. 

Let $\{x_t\}_{t=0}^T$ be the trajectory to be observed by the learner. If $|x_0| > p - 1$, then observe that 

\begin{align*}
\operatorname{FR}_{\mathcal{A}}(T, \mathcal{F}) &= \mathbbm{E}\left[\sum_{t=1}^T \mathbbm{1}\{\mathcal{A}(x_{<t}) \neq x_t\} - \inf_{f \in \mathcal{F}} \sum_{t=1}^T \mathbbm{1}\{f^t(x_0) \neq x_t\} \right]\\
&= \mathbbm{E}\left[\sum_{t=1}^T \mathbbm{1}\{|x_{0}| + t \neq x_t\} - \inf_{f \in \mathcal{F}} \sum_{t=1}^T \mathbbm{1}\{|x_{0}| + t \neq x_t\} \right] = 0\\
\end{align*}

On the other hand, if $|x_0| \leq p - 1$, we can write the learner's expected loss as 

\begin{align*}
\mathbbm{E}\left[\sum_{t=1}^T \mathbbm{1}\{\mathcal{A}(x_{<t}) \neq x_t\}\right] &= \mathbbm{E}\left[\sum_{t=1}^{p - |x_0|} \mathbbm{1}\{\epsilon_t(|x_{0}| + t) \neq x_t\}\right] + \sum_{t=p - |x_0| + 1}^{T} \mathbbm{1}\{|x_0| + t \neq x_t\}.
\end{align*}

Similarly, we can write the cumulative loss of the competitor term as: 

\begin{align*}
\inf_{f \in \mathcal{F}} \sum_{t=1}^T \mathbbm{1}\{f^t(x_0) \neq x_t\} &= \inf_{f \in \mathcal{F}}\sum_{t=1}^{p - |x_0|} \mathbbm{1}\{f^t(x_0) \neq x_t\} + \sum_{t=p - |x_0| + 1}^{T} \mathbbm{1}\{|x_0| + t \neq x_t\}.
\end{align*}

Let $f_{\sigma} = \argmin_{f \in \mathcal{F}}\sum_{t=1}^{p - |x_0|} \mathbbm{1}\{f^t(x_0) \neq x_t\}$. Combining both bounds, we get that:

\begin{align*}
\operatorname{FR}_{\mathcal{A}}(T, \mathcal{F}) &= \mathbbm{E}\left[\sum_{t=1}^T \mathbbm{1}\{\mathcal{A}(x_{<t}) \neq x_t\} - \inf_{f \in \mathcal{F}} \sum_{t=1}^T \mathbbm{1}\{f^t(x_0) \neq x_t\} \right]\\
&\leq \mathbbm{E}\left[\sum_{t=1}^{p - |x_0|} \mathbbm{1}\{\epsilon_t(|x_{0}| + t) \neq x_t\} - \mathbbm{1}\{f_{\sigma}^{t}(x_0) \neq x_t\} \right]\\
&\leq \mathbbm{E}\left[\sum_{t=1}^{p - |x_0|} \mathbbm{1}\{\epsilon_t(|x_{0}| + t) \neq f^{t}_{\sigma}(x_0)\}\right]\\
&= \mathbbm{E}\left[\sum_{t=1}^{p - |x_0|} \mathbbm{1}\{\epsilon_t(|x_{0}| + t) \neq \sigma_{|x_0| + t - 1}(|x_{0}| + t)\}\right] = \frac{p - |x_0|}{2}, 
\end{align*}

\noindent where in the last equality we used the fact that $\epsilon_t \sim \text{Unif}(-1, 1)$.  Thus, the expected regret of such an learner $\mathcal{A}$ is at most $\frac{\min\{\max\{p - |x_0|, 0\}, T\}}{2}$. Taking $x_0 = 0$, gives that $\inf_{\mathcal{A}}\operatorname{FR}_{\mathcal{A}}(T, \mathcal{F}) \leq \frac{\operatorname{C}_T(\mathcal{F})}{2}.$

\subsection{Tightness of upper bound in Theorem \ref{thm:02agn}} \label{exm: agnuptight}
 \noindent Let $\mathcal{X} = \mathbb{Z}$ and fix $p \in \mathbb{N}$. Let $\mathcal{G} = \{(0, s_1, ..., s_{p}):  s_1 < s_2 < ... < s_{p} \in \mathbb{N}\}$ be the set of all ordered tuples of extended natural numbers with size $p+1$. For any $S \in \mathcal{G}$ and $x \in \mathbb{N} \cup \{0\}$, let $S_x = \max\{s \in S: s \leq x\}$ be the largest element in $S$ smaller than $x$. Note that $S_0 = 0$ for all $S \in \mathcal{G}$. For every $\sigma \in \{-1, 1\}^{\mathbb{N} \cup \{0\}}$ and $S \in \mathcal{G}$, consider the evolution function: 

 $$f_{\sigma, S}(x) = \sigma_{S_{|x|}}(|x| + 1).$$

 Given any initial state $x_0 \in \mathcal{X}$ and time horizon $T \in \mathbb{N}$, the trajectory $\{f^{t}_{\sigma, S}(x_0)\}_{t=1}^T$ is some signed version of the sequence $(|x_0| + 1, ..., |x_0| + T)$, where the signs depend on $S$ and $\sigma$. More importantly,  the trajectory $\{f^{t}_{\sigma, S}(x_0)\}_{t=1}^T$ switches signs at most $p+1$ times.  Finally, consider the evolution class $\mathcal{F} = \{f_{\sigma, S}: \sigma \in \{-1, 1\}^{\mathbb{N} \cup \{0\}}, S \in \mathcal{G}\}.$ Since the trajectory of any evolution $f \in \mathcal{F}$ can switch signs at most $p+1$ times, it follows that $\operatorname{C}_T(\mathcal{F}) \leq p+1$. Moreover, by considering a trajectory tree with the root node labeled by $0$ and the set of evolutions parameterized by the tuple $(0, 1, ..., p) \in \mathcal{G}$, its not hard to see that $\operatorname{C}_T(\mathcal{F}) \geq p+1.$ Thus, $\operatorname{C}_T(\mathcal{F}) = p+1$. 
 
 We now claim that $\inf_{\mathcal{A}} \operatorname{FR}_{\mathcal{A}}(T, \mathcal{F}) \geq \sqrt{\frac{\operatorname{C}_T(\mathcal{F}) \, T}{8}}$, which shows that the upper bound in Theorem \ref{thm:02agn} is tight up to a logarithmic factor in $T$ since $\sup_{x \in \mathcal{X}}|\mathcal{F}(x)| = 2$.  Consider $T = k (p+1)$ for some odd $k \in \mathbb{N}$. For $\epsilon \in \{-1, 1\}^T$, define $\tilde{\epsilon}_i = \text{sign}\Bigl(\sum_{t=(i-1)k + 1}^{ik} \epsilon_t \Bigl)$ for all $i \in \{1,2, ..., p+1\}$. The game proceeds as follows. The adversary samples a string $\epsilon \in \{-1, 1\}^T$ uniformly at random and constructs the random trajectory $\{x_t\}_{t=0}^T$ where $x_0 = 0$ is the initial state and $x_t = \epsilon_t t$ for all $t \geq 1$.  The adversary then passes $\{x_t\}_{t=0}^T$ to the learner. 

 Let $\mathcal{A}$ be any randomized learner. Then, for each block $i \in [p+1]$, we have that 

 $$\mathbbm{E}\left[\sum_{t=(i-1)k + 1}^{ik} \mathbbm{1}\{\mathcal{A}(x_{<t}) \neq x_t\} \right] \geq \sum_{t=(i-1)k + 1}^{ik} \frac{1}{2} = \frac{k}{2},$$

 where the inequality follows from the fact that $x_t$ is chosen uniformly at random between $t$ and $-t$. Let $f_{\sigma, S} \in \mathcal{F}$ be the function in $\mathcal{F}$ such that $S = (0, k, 2k, ..., (p+1)k)$ and $\sigma_{S_{|x|}} = \tilde{\epsilon}_i$ for all $(i-1)k \leq |x| \leq ik-1$ and $i \in \{1, ..., p+1\}$. For each block $i \in [p+1]$, we have 

 \begin{align*}
 \mathbbm{E}\left[\sum_{t=(i-1)k + 1}^{ik} \mathbbm{1}\{f^{t}_{\sigma, S}(0) \neq x_t\} \right] &= \mathbbm{E}\left[\sum_{t=(i-1)k + 1}^{ik} \mathbbm{1}\{\sigma_{S_{|t-1|}}t \neq \epsilon_t t\} \right]\\
 &= \mathbbm{E}\left[\sum_{t=(i-1)k + 1}^{ik} \mathbbm{1}\{ \tilde{\epsilon}_i  \neq \epsilon_t \} \right]\\
 &= \frac{k}{2} - \frac{1}{2}\mathbbm{E}\left[\sum_{t=(i-1)k + 1}^{ik}  \tilde{\epsilon}_i \epsilon_t  \right]\\
 &= \frac{k}{2} - \frac{1}{2}\mathbbm{E}\left[\Bigl|\sum_{t=(i-1)k + 1}^{ik} \epsilon_t  \Bigl|\right]\\
 &\leq \frac{k}{2} - \sqrt{\frac{k}{8}},
 \end{align*}

 where the final step follows upon using Khinchine's inequality \citep{cesa2006prediction}. Combining these two bounds above, we obtain, 

 $$\mathbbm{E}\left[\sum_{t=(i-1)k + 1}^{ik} \mathbbm{1}\{\mathcal{A}(x_{<t}) \neq x_t\}  - \sum_{t=(i-1)k + 1}^{ik} \mathbbm{1}\{f^{t}_{\sigma, S}(0) \neq x_t\}\right] \geq \sqrt{\frac{k}{8}}.$$

 Summing this inequality over $p+1$ blocks, we obtain

 \begin{align*}
 \mathbbm{E}\left[\sum_{t=1}^T \mathbbm{1}\{\mathcal{A}(x_{<t}) \neq x_t\}  - \inf_{f \in \mathcal{F}}\sum_{t=1}^{T} \mathbbm{1}\{f^{t}(x_0) \neq x_t\}\right] &\geq  \mathbbm{E}\left[\sum_{t=1}^T \mathbbm{1}\{\mathcal{A}(x_{<t}) \neq x_t\}  -\sum_{t=1}^{T} \mathbbm{1}\{f_{\sigma, S}^{t}(x_0) \neq x_t\}\right] \\
 &\geq (p+1)\sqrt{\frac{k}{8}} = \sqrt{\frac{\operatorname{C}_T(\mathcal{F})T}{8}},
 \end{align*}
 which completes our proof. 

 \section{Proof of Theorem \ref{thm:sepflowregret}} \label{app:proofofsep}
Our proof of Theorem \ref{thm:sepflowregret} is constructive. That is, we provide an evolution function class and show that there exists a deterministic algorithm with a mistake bound of $3$ in the realizable setting. As for the agnostic setting, we use the probabilistic method to argue the existence of a hard stream such that every algorithm incurs $T/6$ regret.

Let $\mathcal{X} = \{-1, 1\}^{\mathbb{N}} \times \mathbb{Z}$.  For every $\sigma \in \{-1, 1\}^{\mathbb{N}}$, define an evolution function 
$$
f_{\sigma}((\theta, z)) = (\sigma, \sigma_{|z|+1}(|z|+1)) \,\,\, \mathbbm{1}\{|z|+1 =0 \, (\text{mod } 3)\} +  (\mathbf{1}, \sigma_{|z|+1}(|z|+1))\,\,\, \mathbbm{1}\{|z|+1  \neq 0 \, (\text{mod } 3)\}, 
$$
where $\theta \in \{-1, 1\}^{\mathbb{N}}, z \in \integers$, and $\mathbf{1} := (1,1,\ldots, 1,1)$ is the string of all ones. Consider the evolution function class $\mathcal{F} = \Bigl\{f_{\sigma}: \sigma \in \{-1, 1\}^{\mathbb{N}}\Bigl\}.$

To prove (i), it suffices to note that for any initial state $(\theta_0, z_0)$ and $f_{\sigma} \in \Fcal$, the realizable stream $\{f_{\sigma}^t((\theta_0, z_0))\}_{t =1}^T$  must reveal $ \sigma$ within the first three rounds $t \in \{1,2,3\}$. A deterministic algorithm $\mathcal{A}$ that plays arbitrarily in the beginning and $f^t_{\sigma}((\theta_0, z_0))$ for all $t \geq 4$ makes no more than $3$ mistakes. 

To prove (ii), consider a random trajectory such that $x_0 = (\mathbf{1}, 0)$ is the initial state and $\{x_t\}_{t=1}^T = \{( \mathbf{1}, \varepsilon_t \, t)\}_{t=1}^T$, where $\varepsilon_t \sim \text{Uniform}(\{-1, 1\})$.  Since the stream is generated uniformly at random, for any  algorithm $\Acal$, we must have 
\[\mathbb{E}\left[\sum_{t=1}^T \indicator\{\Acal(x_{<t}) \neq (\mathbf{1},  \varepsilon_t t) \} \right] \geq \frac{T}{2}.\]

Next, let $\varepsilon \in \{-1,1\}^{\naturals}$ be any completion of $(\varepsilon_1, \ldots, \varepsilon_T)$ and consider the function $f_{\varepsilon} \in \Fcal$. Note that, for every $t \in [T]$, we have $f_{\varepsilon}(\mathbf{1}, \varepsilon_{t-1} (t-1)) = (\mathbf{1},\varepsilon_t t)  $ if $t \neq 0\, (\text{mod } 3)$ and $f_{\varepsilon}(\mathbf{1}, \varepsilon_{t-1} (t-1)) = (\varepsilon, \varepsilon_t \, t) \neq (\mathbf{1},\varepsilon_t t)$ if $t = 0 \text{ (mod)} \, 3$. 
Moreover, it is not too hard to see that 
\[ f^t_{\varepsilon}((\mathbf{1}, 0)) = (\varepsilon, \varepsilon_t t) \,\,\, \mathbbm{1}\{t =0 \, (\text{mod } 3)\} +  (\mathbf{1}, \varepsilon_t t)\,\,\, \mathbbm{1}\{t  \neq 0 \, (\text{mod } 3)\}. \]
Since the functions $f \in \Fcal$ ignore the first argument of the tuple, we have $
f_{\varepsilon}^t((\mathbf{1}, 0)) = f_{\varepsilon}(f_{\varepsilon}^{t-1}(\mathbf{1}, 0)) = f_{\varepsilon}((\mathbf{1}, \varepsilon_{t-1}(t-1))).$ So, using the equality established above, we obtain
\begin{equation*}
    \begin{split}
       \inf_{f \in \mathcal{F}}\sum_{t=1}^T  \indicator\{f^t(x_0) \neq  x_t\} &\leq  \sum_{t=1}^T  \indicator\{f_{\varepsilon}^t(x_0) \neq  x_t\} \\
       &\leq  \sum_{t=1}^T  \indicator\{f_{\varepsilon}^t((\mathbf{1}, 0)) \neq (\mathbf{1}, \varepsilon_{t} t) \} \\
       &= \sum_{t=1}^T \indicator\{f_{\varepsilon}((\mathbf{1}, \varepsilon_{t-1} (t-1))) \neq  (\mathbf{1}, \varepsilon_{t} t)\} \\
       &= \sum_{t=1}^T \indicator\{t = 0 \, (\text{mod } 3)\} \leq \frac{T}{3}.
    \end{split}
\end{equation*}

Therefore, we have $\operatorname{FR}_{\Acal}(T, \mathcal{F}) \geq \frac{T}{2}-\frac{T}{3} \geq \frac{T}{6}.$

\end{document}